\documentclass[runningheads]{llncs}

\usepackage{eccv}

\usepackage{eccvabbrv}

\usepackage{graphicx}
\usepackage{booktabs}

\usepackage[accsupp]{axessibility}  

\usepackage{hyperref}

\usepackage{orcidlink}

\usepackage[dvipsnames]{xcolor}         
\definecolor{linkColor}{rgb}{0.18,0.39,0.62} 

\usepackage{microtype}
\usepackage{graphicx}
\usepackage{booktabs} 
\usepackage{multirow}
\usepackage{pifont}
\usepackage{stmaryrd}
\usepackage{wrapfig}
\usepackage{amsfonts}

\definecolor{brickred}{rgb}{0.8, 0.25, 0.33}
\definecolor{brickgreen}{rgb}{0.25, 0.8, 0.33}
\newcommand{\cm}{\textcolor{brickgreen}{\ding{51}}}%
\newcommand{\xm}{\textcolor{brickred}{\ding{55}}}%
\DeclareMathSymbol{\shortminus}{\mathbin}{AMSa}{"39}

\usepackage{color}
\definecolor{brickred}{rgb}{0.8, 0.25, 0.33}
\definecolor{brickred2}{rgb}{0.25, 0.8, 0.33}
\usepackage{graphicx}

\newcommand{\ttabref}[1]{Tab.~\ref{#1}}
\newcommand{\ffigref}[1]{Fig.~\ref{#1}}
\newcommand{\ssecref}[1]{Sec.~\ref{#1}}
\newcommand{\eeqref}[1]{Eq.~(\ref{#1})}

\newcommand{\framework}{YeTI}

\begin{document}

\title{YeTI: You Only Need Two Noisy Images for Real-World sRGB Noise Generation} 

\titlerunning{YeTI}

\author{
Jaekyun Ko\inst{1,2}$^{*}$ \and
Byung Wan Lim\inst{1}$^{*}$ \and
Soomin Lee\inst{1} \and
Dongjin Kim\inst{1} \and \\
Tae Hyun Kim\inst{1}$^{\dagger}$
}


\authorrunning{J. Ko et al.}

\institute{Department of Computer Science, Hanyang University
\email{\{pook0612,min001017,dongjinkim,taehyunkim\}@hanyang.ac.kr}
\\
\and Mobile Experience (MX) Division, Samsung Electronics\\
\email{jkko1124.ko@samsung.com}}

\maketitle

\begingroup
\renewcommand\thefootnote{}
\footnotetext{* Equal contribution. \quad $\dagger$ Corresponding author.}
\endgroup
\vspace{-8.5mm}

\begin{figure}[h!]
\begin{center}
\vspace{-2.5mm}
\centerline{\includegraphics[width=1.0\textwidth]{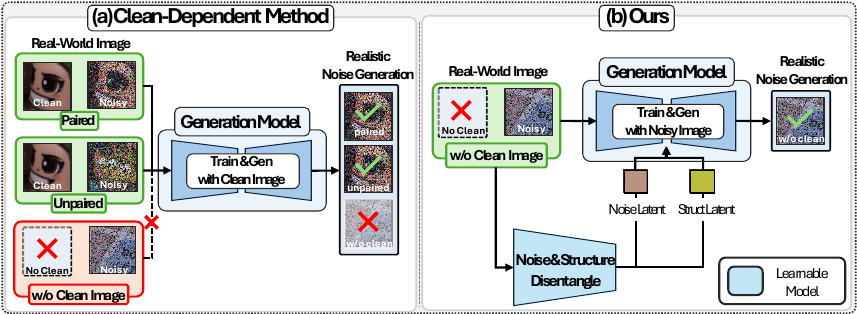}}
\caption{Noise Generation Comparison. (a) Conventional clean-image-dependent approaches. (b) Our clean-image-free approach.}
\vspace{-22.5mm}
\label{fig:teaser}
\end{center}
\end{figure}

\begin{abstract}
Real-world sRGB image denoising remains challenging due to the nonlinear characteristics of sensor noise and the difficulty of acquiring aligned clean-noisy image pairs. Supervised denoisers often overfit to limited paired datasets, while self-supervised methods still depend on sufficiently diverse noisy observations. These limitations motivate scalable noise synthesis methods that can model real-world noise without clean ground truth or camera metadata.
We propose \framework{}, a real-world sRGB noise generation framework that learns from only two noisy observations of the same scene. \framework{} uses a Reconstruction Autoencoder to disentangle scene structure and noise characteristics, and models the latent noise distribution with a one-step Conditional Diffusion Transformer trained using consistency objectives. Given a single noisy input at inference time, \framework{} generates realistic, signal-dependent noise while preserving the underlying scene content.
Extensive experiments demonstrate the effectiveness of \framework{} across real-world benchmarks. We evaluate noise generation on SIDD and further assess generalization on SIDD+, MAI2021, and SID, covering smartphone and diverse consumer-camera sensors. Downstream denoising results on DND further show that denoisers trained with \framework{}-synthesized images achieve strong real-world performance, highlighting the practical value of clean-image-free and metadata-free noise generation.
Code is available at:
\url{https://github.com/ByungWanLim/YeTI-You-Only-Need-Two-Noisy-Images-for-Real-World-sRGB-Noise-Generation}
\end{abstract}

\vspace{-10mm}
\section{Introduction}
\label{sec:intro}
\vspace{-2mm}

Image denoising is a fundamental problem in computer vision and plays a critical role in various real-world applications such as photography~\cite{sid, hdr+}, surveillance~\cite{retinex}, and autonomous driving~\cite{dark_zurich, cityscapes_c, coco_c}. 
Despite the remarkable progress of deep learning-based denoisers~\cite{dncnn, swinir, restormer, nafnet, hinet, uformer, mirnet, mambair}, achieving robust performance in various imaging conditions remains a challenging problem.

In particular, real-world sRGB denoising poses unique difficulties due to the complex characteristics of sensor noise and the nonlinear transformations applied during the image signal processing (ISP) pipeline~\cite{noisepipeline1, noisepipeline2, noisepipeline3}. The raw sensor noise becomes highly distorted after passing through demosaicing, tone mapping, gamma correction, and other camera-specific operations. Consequently, the noise distribution in the real-world sRGB domain deviates significantly from simple Poisson-Gaussian (PG) assumptions~\cite{pg, nfsrgb}, making it difficult for supervised denoising models to generalize well to real-world settings.

Moreover, the scarcity of real-world training datasets leads to overfitting problems in supervised denoising methods.
Although datasets such as SIDD~\cite{sidd} and MIDD~\cite{midd} attempt to collect real-world noisy–clean image pairs, the process is labor-intensive and expensive. 
This process requires capturing more than a thousand short- and long-exposure images for each scene under fixed camera settings and controlled environmental conditions, followed by additional post-processing to construct high-quality ground truth images.

\begin{table}[t]
    \centering
    \caption{Comparison between real-world sRGB noise modeling methods.}
    \vspace{-1mm}
    \resizebox{0.8\columnwidth}{!}{
    \begin{tabular}{l|ccc}
    \toprule[0.5pt]
    Generative Model & No Need for Clean? & Realistic Noise? & No Need for Metadata? \\ \midrule[0.2pt]
    (a) NeCA, NAFlow   & \xm & \cm & \xm \\
    (b) C2N                      & \xm & \xm & \cm \\
    (c) SeNM-VAE    & \xm & \cm & \cm \\
    \textbf{(d) \framework{} (Ours)} & \cm & \cm & \cm \\ \bottomrule[0.5pt]
    \end{tabular}
    }
    \label{tab:cmp_method}
    \vspace{-5mm}
\end{table}

To reduce the dependence on large-scale real-world data collection, recent studies~\cite{c2n, naflow, noiseflow, nfsrgb, lee2022noisetransfer, neca, senm_vae} have modeled real-world noise distributions to synthesize additional noisy samples. Nevertheless, many of these methods require paired clean-noisy images or hardware-specific metadata, such as ISO and shutter speed, which are often unavailable in practical scenarios.

To address these limitations, we introduce YeTI, a clean-image-free framework for real-world sRGB noise generation. As illustrated in \ffigref{fig:teaser} and summarized in \ttabref{tab:cmp_method}, our framework eliminates the need for paired clean-noisy images and camera metadata while synthesizing realistic sRGB noise.
Existing sRGB noise modeling methods differ substantially in their data requirements and generation realism.
Prior methods such as NeCA~\cite{neca} and NAFlow~\cite{naflow} can synthesize realistic noise, but require either camera metadata or paired clean-noisy images for training or noise synthesis. Although C2N~\cite{c2n} learns noise generation in an unpaired setting, producing realistic noise remains challenging due to the instability of GAN-based unpaired training~\cite{wgan, improved_gan}. SeNM-VAE~\cite{senm_vae} improves the realism of generated noise without relying on metadata, but still depends on paired datasets during training. In contrast, \framework{} synthesizes realistic sRGB noise using only two noisy burst observations during training and a single noisy image during inference, making it substantially more practical for real-world data acquisition.

At the core of our approach is a Reconstruction Autoencoder (RAE) that learns a compact latent space for disentangling scene structure and noise characteristics from input noisy images using only two consecutively captured burst observations. The RAE consists of two specialized encoders, namely a Structure Encoder and a Noise Encoder, together with a shared decoder. The Structure Encoder is trained with a contrastive objective~\cite{infonce} to extract noise-invariant content representations, where the two burst noisy images from the same scene form a positive pair and images from other scenes within the mini-batch serve as negatives. Conditioned on these structural representations, the Noise Encoder captures residual noise-specific information in a dedicated noise latent. The shared decoder then reconstructs the noisy input using both structural and noise latent features, preserving the underlying scene content while retaining signal-dependent and spatially correlated noise characteristics.

To model the distribution of the noise latent, we train a Conditional Diffusion Transformer (C-DiT) with a one-step consistency objective~\cite{cm,latentcm,ict,scm}. The C-DiT is conditioned on the structure and noise latent features extracted from one noisy observation in the burst pair, and learns to predict the noise latent extracted from the other observation. This conditional training strategy provides sensor-specific noise guidance without requiring clean images or camera metadata, enabling the model to synthesize novel noise samples rather than merely reproducing a particular noise instance.

During inference, we sample a latent code from a simple prior and transform it with C-DiT into a noise latent conditioned on the structure and noise representations extracted from a single noisy image. The RAE decoder then maps the generated latent back to the image space, producing a synthesized noisy image that preserves the underlying scene structure while exhibiting realistic noise characteristics.

Through extensive experiments, we demonstrate that \framework{} achieves state-of-the-art noise modeling performance across diverse real-world sensor domains. Furthermore, downstream denoising experiments show that denoisers trained with \framework{}-synthesized noisy images achieve strong real-world performance, confirming the practical value of clean-image-free and metadata-free noise generation.

\vspace{-4mm}
\section{Related Works}
\label{related_works}
\vspace{-2mm}
\subsection{Real-World Image Denoising}
\vspace{-2mm}
Image denoising has long been a fundamental problem in low-level vision and image processing. With the rapid progress of deep learning, recent approaches have largely shifted toward data-driven frameworks that learn complex noise-to-clean mappings from large-scale training data. Existing denoising methods can be broadly grouped into supervised and self-supervised paradigms.

Supervised methods train denoisers using paired clean–noisy images~\cite{dncnn, ircnn, ffdnet, drunet, swinir, restormer, nafnet, hinet, uformer, mirnet, mambair, mambair_v2, idf}. Representative models, including NAFNet~\cite{nafnet} and MambaIR~\cite{mambair}, adopt expressive architectures based on attention mechanisms or state-space modeling to enlarge the receptive field and capture complex degradation patterns. Despite their strong restoration performance, these methods often depend heavily on the training noise distribution, which can lead to degraded performance under unseen camera devices or imaging conditions. Such sensitivity limits their generalization ability in diverse real-world environments.

Self-supervised methods alleviate the need for clean ground truth images by exploiting internal noise statistics or spatial redundancy~\cite{apbsn, lgbpn, 2020_bsn, mm_bsn, bsn_2018, bsn_2019}. Blind-spot-based approaches, such as AP-BSN~\cite{apbsn} and LG-BPN~\cite{lgbpn}, have shown promising performance on real-world noisy images by breaking spatial noise correlations during training. However, their effectiveness still largely depends on the diversity and representativeness of the available noisy data. When the noisy observations are limited or biased toward specific imaging conditions, self-supervised denoisers may overfit or fail to capture the full real-world noise distribution. As a result, both supervised and self-supervised paradigms remain constrained by the difficulty of acquiring diverse, high-quality real-world denoising datasets.

\vspace{-4.5mm}
\subsection{Real-World sRGB Noise Generation}
\vspace{-2mm}
To mitigate overfitting caused by limited real-world paired datasets, recent studies have explored sRGB noise modeling methods that synthesize realistic noisy images. These methods aim to capture the statistical, spatial, and signal-dependent characteristics of camera sensor noise, thereby enabling more robust denoiser training without requiring large-scale additional data collection.

NeCA~\cite{neca} introduces a neighboring correlation-aware model that explicitly considers signal dependency and local noise correlations to synthesize realistic noise. NAFlow~\cite{naflow} further models sRGB noise distributions using a noise-aware normalizing flow, enabling realistic noise sampling without camera-specific metadata during inference. C2N~\cite{c2n} formulates real-world noise generation as an unpaired image-to-image translation problem, removing the need for paired supervision and metadata. SeNM-VAE~\cite{senm_vae} adopts a variational inference framework that supports paired and partially unpaired training schemes. Despite these advances, existing methods still face practical limitations. NeCA, NAFlow, and SeNM-VAE require either metadata or paired clean-noisy images during training, while C2N often struggles to faithfully model real-world sRGB noise due to the instability of GAN-based unpaired learning. These requirements limit their practicality and scalability in real-world data acquisition scenarios.

In contrast, our method relaxes these constraints by learning from only two noisy burst images of the same scene. This design removes the dependency on clean references and camera metadata during both training and inference, enabling scalable, metadata-free, and realistic sRGB noise synthesis.

\vspace{-3.5mm}
\section{Proposed Method}
\vspace{-2mm}
\subsection{Preliminaries: One-step Diffusion}
\label{sec:3.1}

\noindent\textbf{Diffusion Models.}
Diffusion models~\cite{sohdiffusion, songscore1, hodiffusion} generate data by reversing a forward corruption process in which \textit{i.i.d.} Gaussian noise is added to a clean sample $\mathbf{x}_0$. Using reparameterization, the noisy sample at timestep $t$ is written as:
\vspace{-1mm}
\begin{equation} 
\mathbf{x}_t = \alpha_t \mathbf{x}_0 + \sigma_t \pmb{\epsilon}, \quad \pmb{\epsilon} \sim \mathcal{N}(0, \pmb{\mathcal{I}}), \quad t  \in \{0, 1, \dots, T\},
\label{eq:1}
\end{equation}
where $\pmb{\epsilon}$ denotes Gaussian random noise, $\sigma_t$ controls the noise level and we set $\alpha_t = 1$ following the formulation of EDM~\cite{edm}.

\begin{figure*}[t]
\begin{center}
\centerline{\includegraphics[width=1.0\textwidth]{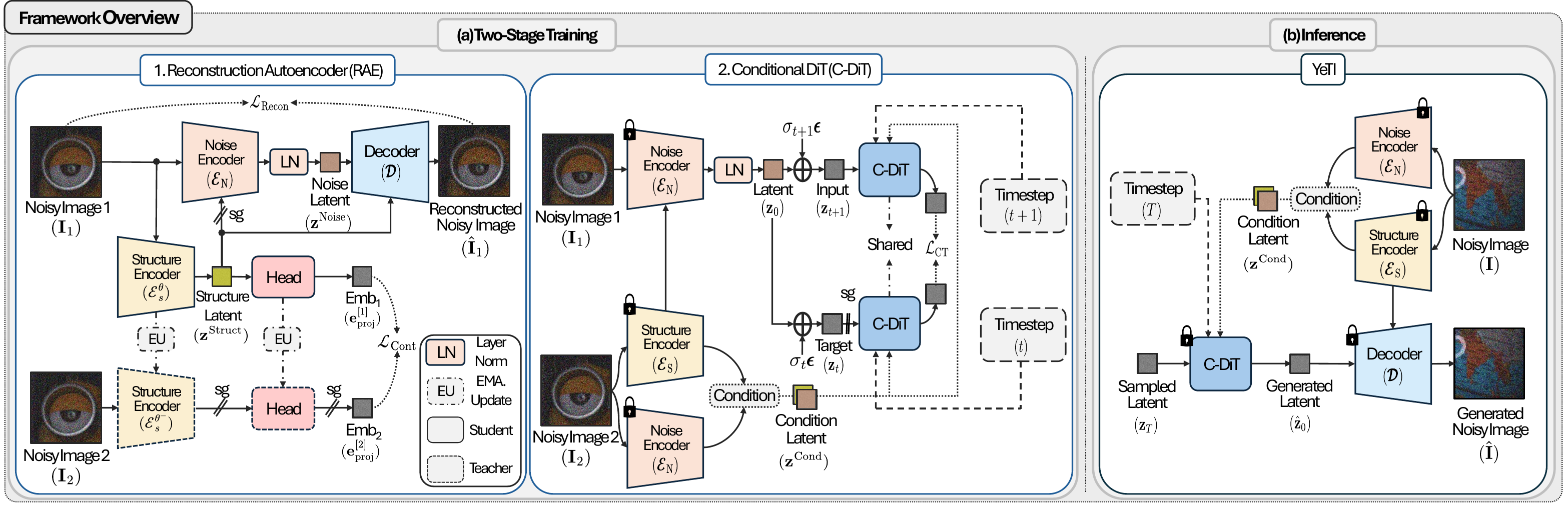}}
\caption{Overview of the proposed method. (a) Two-stage training phase. (b) Inference phase.}
\vspace{-13mm}
\label{fig:overall_flow}
\end{center}
\end{figure*}

The reverse process is learned by solving the probability flow ODE~\cite{edm},
which transports $\mathbf{x}_T$ toward $\mathbf{x}_0$.
Solving this ODE typically requires many update steps, which makes the sampling process computationally demanding.

\noindent\textbf{Consistency Models (CM).}
To enable faster generation, consistency models~\cite{cm, ict} learn a direct single-step mapping that produces predictions consistent with the clean sample $x_0$ across different noise levels $\sigma_t$. Specifically, a consistency model predicts:
\vspace{-1mm}
\begin{equation}
f^\theta(\mathbf{x}_t, \sigma_t) = \mathbf{x}_t + \int_{\sigma_t}^{\sigma_0} \frac{d\mathbf{x}_u}{du} \,du \approx \mathbf{x}_0,
\end{equation}
and is trained with the consistency loss as:
\vspace{-1mm}
{\small
\begin{equation}
\mathcal{L}_\text{CT} = \mathbb{E} \left[ \lambda(\sigma_t) \, d \left( f^\theta(\mathbf{x}_{t+1}, \sigma_{t+1}) - \text{sg}(f^{\theta^{-}}(\mathbf{x}_{t}, \sigma_{t})) \right) \right],
\label{eq:ct_loss}
\end{equation}
}
where $d(\cdot)$ is a distance metric, $\lambda(\cdot)$ is a weighting function, and the stop-gradient operator (\text{sg}) stabilizes the training.
Specifically, the teacher model $f^{\theta^{-}}$ provides a stable distillation target to ensure the student network $f^{\theta}$ learns consistent mappings along the probability flow ODE trajectory.
Following prior work~\cite{cm, ict, actdiffusion}, we do not apply exponential moving average (EMA) updates from the student to the teacher, and instead share parameters between them such that $\theta^{-} = \theta$.
The hyperparameters follow the setting in EDM~\cite{edm} and iCT~\cite{ict}, and additional details are provided in the Supplementary Material Sec. S1. 

\vspace{-3.5mm}
\subsection{Overall Flow}
\label{sec:3.2}
\vspace{-2mm}
In this work, we propose \framework{} to model real-world noise distributions from only a single noisy input during inference. To achieve this, the model must disentangle noise characteristics from the underlying scene structure in a given noisy image. This task is highly challenging, and existing approaches often rely on external information, such as camera metadata or paired clean images, to guide the separation.

\begin{figure*}[ht]
\begin{center}
\centerline{\includegraphics[width=1.0\textwidth]
{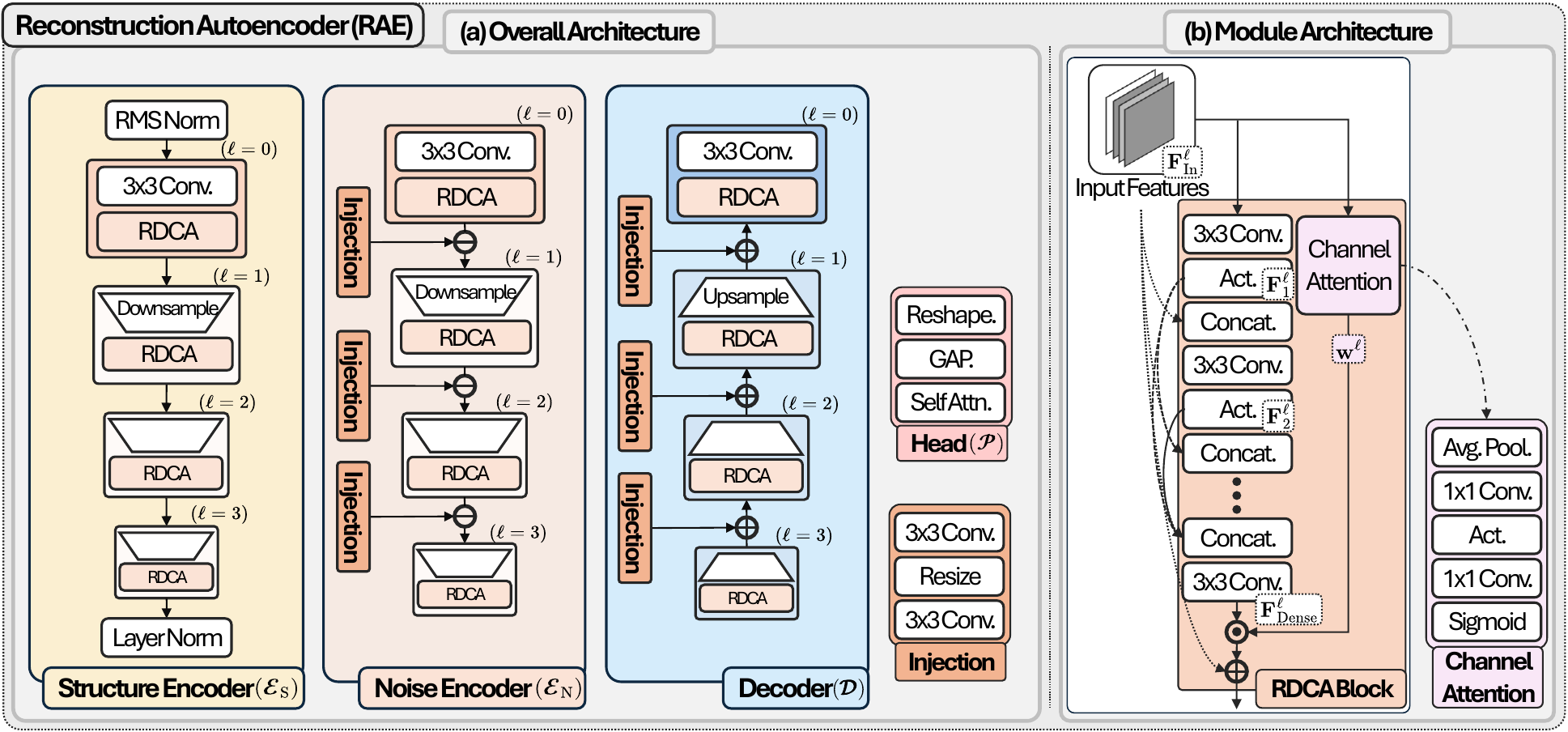}}
\caption{Detailed configuration. (a) Reconstruction Autoencoder (RAE). (b) Residual Dense Channel Attention (RDCA) block.}
\vspace{-12.5mm}
\label{fig:recon_ae}
\end{center}
\end{figure*}

In contrast, our approach learns to disentangle noise characteristics from the underlying scene structure by using burst noisy images captured from the same scene under identical camera settings during training. This design exploits the signal consistency shared across burst observations, allowing transient noise components to be isolated without clean references or camera metadata. 

As illustrated in \ffigref{fig:overall_flow}(a), \framework{} consists of two main components: a Reconstruction Autoencoder (RAE) and a Conditional Diffusion Transformer (C-DiT). Following common latent diffusion practice~\cite{ldm}, the framework is trained in two stages. First, the RAE learns a disentangled latent space that separates scene structure from stochastic noise using burst noisy observations. Then, with the RAE fixed, the C-DiT models the distribution of the noise latent through a one-step diffusion process, following the consistency training framework~\cite{latentcm}. During inference, C-DiT transforms a randomly initialized latent into a realistic noise latent conditioned on a single noisy image, which is subsequently decoded by the RAE to synthesize a new noisy image while preserving the original scene content.

\vspace{-5mm}
\subsection{Reconstruction Autoencoder (RAE)}
\label{RAE}
\vspace{-2mm}
As shown in \ffigref{fig:recon_ae}, the key challenge of clean-image-free noise modeling is to disentangle scene structure from stochastic noise using only noisy observations. To address this challenge, we propose a Reconstruction Autoencoder (RAE) consisting of a Structure Encoder $\mathcal{E}_s$, a Noise Encoder $\mathcal{E}_n$, and a Decoder $\mathcal{D}$. 

Unlike previous approaches that rely on synthetic augmentations or restrictive prior assumptions~\cite{fdn}, our framework exploits the physical redundancy of burst noisy images. Since two burst observations captured from the same static scene share identical scene content while containing independent noise realizations, the RAE can naturally separate invariant structural information from transient noise components. Consequently, the learned latent representations better reflect real-world sensor characteristics and provide an effective representation space for subsequent noise modeling.

The RAE first extracts scene-invariant structural representations, then suppresses redundant structural information to obtain a compact noise representation, and finally reconstructs the noisy image by combining both latent representations.

\vspace{-5mm}
\subsubsection{Structure Encoder ($\mathcal{E}_s$).}

The Structure Encoder $\mathcal{E}_s$ serves as the foundation of the RAE by extracting scene-invariant representations shared across burst observations. Let $\mathbf{I}_1$ and $\mathbf{I}_2$ denote two burst noisy images captured from the same static scene, where the subscripts indicate the burst indices. Although $\mathbf{I}_1$ and $\mathbf{I}_2$ contain independent noise realizations, they share identical underlying scene content. We exploit this property through contrastive learning, encouraging representations extracted from the same scene to be aligned while separating those from different scenes. As a result, the $\mathcal{E}_s$ preserves structural information while suppressing noise-specific variations.

As shown in \ffigref{fig:recon_ae}(a), $\mathcal{E}_s$ first applies RMSNorm~\cite{rmsnorm,idf} to stabilize the feature distribution across diverse noise conditions. Multi-scale features are then extracted using Residual Dense Channel Attention (RDCA) blocks, producing the structure latent representations $\mathbf{z}^{\mathrm{Struct}}_1$ and $\mathbf{z}^{\mathrm{Struct}}_2$.
The extracted structure latents are subsequently shared with both the $\mathcal{E}_n$ and the $\mathcal{D}$. Specifically, they are used to suppress redundant scene information during noise encoding and restore structural information during image reconstruction, enabling explicit structure-noise disentanglement throughout the RAE.

\vspace{-5mm}
\subsubsection{Noise Encoder ($\mathcal{E}_n$).}
We introduce a Noise Encoder $\mathcal{E}_n$ to extract noise-specific information from $\mathbf{I}_1$ while suppressing scene structure. As in $\mathcal{E}_s$, $\mathcal{E}_n$ follows the multi-scale design and employs a stack of RDCA blocks to maintain feature compatibility across spatial resolutions. To encourage noise-focused representations, we remove structure-related information from $\mathcal{E}_n$ through subtractive injection at each scale, and later restore it in the decoder through additive injection. Unlike simple feature concatenation, this complementary injection strategy encourages the $\mathcal{E}_n$ to focus on stochastic noise characteristics while allowing the Decoder to recover the removed structural information, leading to improved structure-noise disentanglement.

Specifically, the structure latent $\mathbf{z}^{\mathrm{Struct}}_{1}$ extracted by $\mathcal{E}_{s}$ is transformed into a scale-specific structure-guidance feature:
\vspace{-1mm}
\begin{equation}
\mathbf{G}_{\ell} = \mathrm{Conv}_{3\times3} \left( \mathrm{Resize}_{\ell} \left( \mathrm{Conv}_{3\times3} \left( \mathbf{z}^{\mathrm{Struct}}_{1} \right) \right) \right),
\end{equation}
where $\mathrm{Resize}_{\ell}$ denotes spatial interpolation used to match the feature resolution at scale level $\ell  \in \{1,2,3\}$. At each scale, $\mathbf{G}_{\ell}$ is subtracted from the $\mathcal{E}_n$ features to suppress structural information and isolate stochastic noise residuals. Finally, Layer Normalization ($\mathrm{LN}$)~\cite{ln} is applied to stabilize the noise feature distribution across diverse imaging conditions, improving optimization stability and producing the noise latent $\mathbf{z}^{\mathrm{Noise}}_1$.

\vspace{-4mm}
\subsubsection{Decoder ($\mathcal{D}$).}
The Decoder $\mathcal{D}$ adopts a symmetric multi-scale architecture similar to the two encoders. Given the noise latent $\mathbf{z}^{\mathrm{Noise}}_{1}$ as input, it progressively reconstructs the noisy image through a multi-scale decoding process. Starting from the coarsest level (\ie, $\ell = 3$), it progressively upsamples the features and refines them with RDCA blocks. 
Since structural information is intentionally suppressed in $\mathcal{E}_n$, $\mathcal{D}$ restores it through additive injection at each scale using the corresponding structure-guidance feature $\mathbf{G}_{\ell}$. This complementary design preserves scene fidelity while allowing $\mathcal{E}_n$ to remain dedicated to modeling stochastic noise.

\vspace{-4mm}
\subsubsection{Residual Dense Channel Attention (RDCA) Block.}
To encode scene structure and noise statistics into compact latent representations, we introduce a Residual Dense Channel Attention (RDCA) block. The RDCA block captures local spatial dependencies through dense convolutional connections and global channel relationships through channel attention.

As shown in~\ffigref{fig:recon_ae} (b), the RDCA block first extracts dense spatial features using a stack of convolutional layers~\cite{densenet}. Let $\mathbf{F}^{\ell}_{\mathrm{In}}\in\mathbb{R}^{\frac{H}{2^\ell}\times\frac{W}{2^\ell}\times C^\ell}$ denote the input feature at scale level $\ell \in \{0,1,2,3\}$, where $H$ and $W$ are the spatial resolution of the input image and $C^\ell$ is the number of channels at scale $\ell$. The dense feature embedding is defined as:
\vspace{-1mm}
\begin{equation}
\mathbf{F}^{\ell}_{\mathrm{Dense}} = \phi\left([\mathbf{F}^{\ell}_{1}, \mathbf{F}^{\ell}_{2}, \ldots, \mathbf{F}^{\ell}_{K}]\right),
\end{equation}
where $\mathbf{F}^{\ell}_{i} = \delta(\mathrm{Conv}_{3\times 3}(\mathbf{F}^{\ell}_{i-1}))$ denotes the $i$-th intermediate feature, $\delta(\cdot)$ is a LeakyReLU activation, $[\cdot]$ indicates channel-wise concatenation, and $\phi(\cdot)$ is a $3\times3$ convolution used for feature refinement.

To incorporate global contextual information, we apply channel attention~\cite{channel_attention, nafnet, restormer} to the dense feature:
\vspace{-1mm}
\begin{equation}
\mathbf{F}^{\ell}_{\mathrm{Attn}} = \mathbf{w}^{\ell} \odot \mathbf{F}^{\ell}_{\mathrm{Dense}},
\end{equation}
where $\odot$ denotes element-wise multiplication. The channel-wise attention weight $\mathbf{w}^{\ell}$ is computed as:
\vspace{-1mm}
\begin{equation}
\mathbf{w}^{\ell} = \sigma\left(\mathrm{Conv}_{1\times 1}\left(\delta\left(\mathrm{Conv}_{1\times 1}(\mathrm{GAP}(\mathbf{F}^{\ell}_{\mathrm{In}}))\right)\right)\right),
\end{equation}
where $\mathrm{GAP}$ denotes global average pooling and $\sigma$ is the sigmoid function. Finally, a residual connection is used to stabilize training and produce the output feature:
\vspace{-1mm}
\begin{equation}
\mathbf{F}^{\ell}_{\mathrm{Out}} = \mathbf{F}^{\ell}_{\mathrm{In}} \oplus \mathbf{F}^{\ell}_{\mathrm{Attn}},
\end{equation}
where $\oplus$ denotes element-wise addition. By combining dense spatial feature extraction with channel attention, the RDCA block effectively captures both local and global contextual information.

\vspace{-5mm}
\subsubsection{Training Pipeline.}
The RAE is optimized with a multi-task objective that reconstructs the input noisy image while encouraging the latent space to separate scene structure from noise characteristics. The main supervision is the reconstruction loss $\mathcal{L}_{\mathrm{Recon}}$, defined as the Mean Absolute Error (MAE) between the reconstructed image $\hat{\mathbf{I}}_1$ and the input noisy image $\mathbf{I}_1$. This loss preserves pixel-level fidelity by recovering both the underlying signal and its corresponding noise realization.

To learn noise-invariant structural representations, we employ a contrastive loss $\mathcal{L}_{\mathrm{Cont}}$ based on InfoNCE~\cite{infonce}. Specifically, we compute two embeddings, $\mathbf{u}_1 = \mathcal{P}(\mathcal{E}_s(\mathbf{I}_1))$ and $\mathbf{u}_2 = \mathcal{P}(\mathcal{E}_s(\mathbf{I}_2))$, where $\mathcal{P}(\cdot)$ denotes a projection head applied to the output of the $\mathcal{E}_s$. Since $\mathbf{I}_1$ and $\mathbf{I}_2$ are captured from the same scene, $(\mathbf{u}_1, \mathbf{u}_2)$ forms a positive pair, while embeddings from other samples in the mini-batch are treated as negatives. This objective pulls representations of the same scene closer together despite different noise realizations, while pushing apart representations from different scenes. As a result, $\mathcal{E}_s$ is encouraged to capture structural information rather than noise-specific variations.
To stabilize contrastive learning, we adopt a momentum teacher encoder $\mathcal{E}_s^{\theta^-}$ updated by an exponential moving average (EMA) of the online $\mathcal{E}_s^{\theta}$ and projection head, following common practice in self-supervised representation learning~\cite{byol,moco}. This momentum update provides slowly evolving target representations, facilitating the learning of scene-invariant structural representations from burst observations with different noise realizations. During training, $\mathcal{E}_s^{\theta}$ is optimized by backpropagation, while $\mathcal{E}_s^{\theta^-}$ provides target representations. As illustrated in \ffigref{fig:overall_flow}(a), the stop-gradient (sg) operation prevents gradients from propagating through the teacher branch.

In addition, we apply a total variation loss $\mathcal{L}_{\mathrm{TV}}$~\cite{tvloss} to the structure latent $\mathbf{z}^{\mathrm{Struct}}_{1}$ to promote spatial coherence and suppress residual noise. This regularization helps obtain a cleaner structure representation while preserving the underlying scene content. We also apply an $\mathcal{L}_2$ regularization loss $\mathcal{L}_{\mathrm{Norm}}$ to the noise latent $\mathbf{z}^{\mathrm{Noise}}_{1}$ extracted by the $\mathcal{E}_n$, which constrains the latent scale and improves optimization stability.

The final RAE objective is defined as:
\vspace{-1mm}
\begin{equation}
\label{eq:rae_loss_function}
\mathcal{L}_{\mathrm{RAE}}
=\mathcal{L}_{\mathrm{Recon}}
+\lambda_{\mathrm{Cont}}\mathcal{L}_{\mathrm{Cont}}
+\lambda_{\mathrm{TV}}\mathcal{L}_{\mathrm{TV}}
+\lambda_{\mathrm{Norm}}\mathcal{L}_{\mathrm{Norm}},
\end{equation}
where $\lambda_{\mathrm{Cont}}$, $\lambda_{\mathrm{TV}}$, and $\lambda_{\mathrm{Norm}}$ control the contribution of each loss term. 
This objective jointly optimizes faithful reconstruction, robust structure-noise disentanglement, and stable latent representations, providing an effective representation space for the subsequent diffusion-based noise generation.

\vspace{-4mm}
\subsection{C-DiT: Learning Real-World Noise Distribution}
\vspace{-2mm}

Using the compact latent space learned by the RAE, we model the real-world noise distribution directly in the latent domain. Since the $\mathcal{E}_n$ extracts representations that capture the noise characteristics of the input image, learning a generative model in this space enables realistic noise synthesis.

To this end, we adopt a one-step diffusion framework based on consistency models, inspired by~\cite{png}. This framework learns a direct mapping from a sampled latent to its target noise latent, enabling efficient generation without iterative sampling.

\vspace{-5mm}
\subsubsection{Conditioning and Target Latent Features.}
Given a noisy observation $\mathbf{I}_1$, we first extract the target noise latent using the pretrained $\mathcal{E}_n$. Specifically, $\mathbf{I}_1$ is mapped to
\vspace{-1mm}
\begin{equation}
\mathbf{z}_0 = \mathrm{LN}(\mathcal{E}_n(\mathbf{I}_1)),
\end{equation}
which serves as the target latent for the diffusion model. Layer Normalization ($\mathrm{LN}$) is applied to align the latent distribution and improve the stability of diffusion training.

To guide noise synthesis, we construct a conditioning feature $\mathbf{z}^{\mathrm{Cond}}$ from the other noisy observation $\mathbf{I}_2$ in the burst pair. This condition combines the noise latent ${\mathbf{z}}^{\mathrm{Noise}}_2 = \mathcal{E}_n(\mathbf{I}_2)$ and the structure latent $\mathbf{z}^{\mathrm{Struct}}_2 = \mathcal{E}_s(\mathbf{I}_2)$ through channel-wise concatenation:
\vspace{-1mm}
\begin{equation}
\mathbf{z}^{\mathrm{Cond}} = \left[ {\mathbf{z}}^{\mathrm{Noise}}_2, \mathbf{z}^{\mathrm{Struct}}_2 \right].
\end{equation}
Here, ${\mathbf{z}}^{\mathrm{Noise}}_2$ is used without $\mathrm{LN}$ to preserve raw noise-related statistics, such as variations induced by sensor type and ISO. Meanwhile, $\mathbf{z}^{\mathrm{Struct}}_2$ provides a structural anchor for the underlying scene content. By conditioning on both noise and structure information, C-DiT learns the signal-dependent characteristics of real-world noise and synthesizes realistic noise latent features.

\vspace{-5mm}
\subsubsection{Training Pipeline.}
We apply the diffusion process described in \ssecref{sec:3.1} to the target latent $\mathbf{z}_0$, producing noisy latent variables $\mathbf{z}_t$ for $t \in \{0, 1, \dots, T\}$. At each timestep $t$, the forward process is defined as:
\vspace{-1mm}
\begin{equation}
\mathbf{z}_{t} = \mathbf{z}_{0} \oplus \sigma_{t}\pmb{\epsilon}, \qquad \pmb{\epsilon} \sim \mathcal{N}(0, \pmb{\mathcal{I}}),
\end{equation}
where $\sigma_{t}$ denotes the noise level. Following the consistency objective in \eeqref{eq:ct_loss}, C-DiT is trained to map different points along the same probability flow ODE trajectory to a consistent target state:
\vspace{-1mm}
\begin{equation}
\mathrm{C\text{-}DiT}_{\theta}(\mathbf{z}_{t+1}, \sigma_{t+1} \mid \mathbf{z}^{\mathrm{Cond}})
\approx
\mathrm{C\text{-}DiT}_{\theta}(\mathbf{z}_{t}, \sigma_{t} \mid \mathbf{z}^{\mathrm{Cond}})
\approx
\mathbf{z}_{0}.
\end{equation}
This training objective encourages C-DiT to serve as a stable single-step generation operator across different noise levels.

\vspace{-4mm}
\subsection{sRGB Noise Generation with Single Noisy Image}
\vspace{-2mm}
With the fully trained RAE and C-DiT, \framework{} synthesizes burst noisy images from a single noisy input. As illustrated in~\ffigref{fig:overall_flow} (b), we first sample a random latent $\mathbf{z}_T$ from a predefined normal distribution. Conditioned on latent features extracted from the input noisy image $\mathbf{I}$, C-DiT transforms $\mathbf{z}_T$ into a latent code $\hat{\mathbf{z}}_0$ that captures realistic, signal-dependent noise characteristics. The decoder $\mathcal{D}$ then maps $\hat{\mathbf{z}}_0$ back to the image space to generate a synthesized noisy image $\hat{\mathbf{I}}$. This process enables \framework{} to generate diverse noisy images from a single input while preserving the underlying scene content and matching the input noise distribution.

\vspace{-4mm}
\section{Experiments}
\vspace{-2mm}
\subsection{Experimental Setup}
\label{sec:4.1}
\vspace{-2mm}
\subsubsection{Implementation Details.}
The RAE is trained with the Adam optimizer~\cite{adam} by minimizing the joint objective $\mathcal{L}_{\mathrm{RAE}}$ defined in \eeqref{eq:rae_loss_function}. The weights of the individual loss terms are provided in the Supplementary Material Sec. S2. We initialize the learning rate to $1{\times}10^{-4}$ and use a cosine annealing scheduler~\cite{cosine_annealing} to gradually decrease it to $1{\times}10^{-6}$ over 200k iterations. During training, we use randomly cropped $256{\times}256$ patches with a mini-batch size of 64.

The C-DiT model is trained from scratch without distillation. We optimize it with the RAdam optimizer~\cite{radam} using a fixed learning rate of $2{\times}10^{-4}$ for 100k iterations. Following iCT~\cite{ict}, we use the pseudo-Huber loss to optimize the consistency objective in \eeqref{eq:ct_loss}. The input images are randomly cropped to $256{\times}256$, producing latent codes of size $32{\times}32$, and the mini-batch size is set to 64.

For downstream denoising with the synthesized data, we adopt blind-spot networks (BSNs), which are trained using only noisy images and therefore align well with our noise generation setting. Specifically, we use AP-BSN~\cite{apbsn} and MM-BSN~\cite{mm_bsn} with their official training configurations.

All \framework{} models are trained on four NVIDIA RTX A6000 GPUs and inference is evaluated on a single A6000 GPU.

\vspace{-5mm}
\subsubsection{Metrics.}
We evaluate the quality of the generated noise using the Kullback-Leibler divergence (KLD) and the Average KLD (AKLD)~\cite{danet}. For denoising performance, we report PSNR and SSIM~\cite{ssim} values.

\vspace{-4mm}
\subsubsection{Dataset.}
We train \framework{} on the SIDD training set~\cite{sidd}, which covers 34 camera configurations. Following prior works~\cite{naflow, senm_vae}, we use the SIDD Medium split, which consists of 320 noisy-clean image pairs captured by five smartphones: Google Pixel (GP), iPhone 7 (IP), Samsung Galaxy S6 Edge (S6), Motorola Nexus 6 (N6), and LG G4 (G4). Although clean images are provided, \framework{} does not use them for training; instead, it only exploits the two noisy burst observations available for each scene, which naturally satisfy the input requirement of our framework. We use the SIDD validation set to evaluate both generated noise quality and downstream denoising performance. To assess cross-dataset robustness, we further evaluate noise generation on SIDD+~\cite{siddplus}, MAI2021~\cite{mai2021}, and the See-in-the-Dark (SID) dataset~\cite{sid}, which includes images captured by Fujifilm X-T2 and Sony $\alpha$7S II cameras. Finally, we use the DND benchmark~\cite{dnd} to evaluate downstream denoising performance on unseen real-world domains.

\begin{table}[!h]
    \centering
    \vspace{-0.6cm}
    \caption{Quantitative results of synthetic noise on the SIDD validation subset across different camera devices. The results are computed with KLD$\downarrow$ and AKLD$\downarrow$. The best and second-best results are shown in \textbf{bold} and \underline{underline}, respectively.}
    \vspace{-4mm}
    \resizebox{1.0\linewidth}{!}{
    \begin{tabular}{l|cc|cc|cc|cc|cc|cc}
    \toprule[0.5pt]
    {Camera} & \multicolumn{2}{c|}{G4} & \multicolumn{2}{c|}{GP} & \multicolumn{2}{c|}{IP} & \multicolumn{2}{c|}{N6} & \multicolumn{2}{c|}{S6} & \multicolumn{2}{c}{\textbf{Average}} \\ \cmidrule(lr){1-1} \cmidrule(lr){2-3} \cmidrule(lr){4-5} \cmidrule(lr){6-7} \cmidrule(lr){8-9} \cmidrule(lr){10-11} \cmidrule(lr){12-13}
    {Methods} & KLD$\downarrow$ & AKLD$\downarrow$ & KLD$\downarrow$ & AKLD$\downarrow$ &
KLD$\downarrow$ & AKLD$\downarrow$ &
KLD$\downarrow$ & AKLD$\downarrow$ &
KLD$\downarrow$ & AKLD$\downarrow$ &
KLD$\downarrow$ & AKLD$\downarrow$ \\ \midrule[0.2pt]
    NAFlow (100\%)    & \underline{0.0254} & \underline{0.1367} & \underline{0.0352} & \underline{0.1180} & \underline{0.0339} & 0.1522 & \underline{0.0309} & \underline{0.1108} & \textbf{0.0272} & \underline{0.1355} & \underline{0.0305} & 0.1306 \\
    SeNM-VAE (0.01\%) & 0.0942 & 0.1509 & 0.0490 & 0.1204 & 0.0687 & \textbf{0.1169} & 0.0731 & 0.1244 & 0.0512 & 0.1350 & 0.0672 & \underline{0.1295} \\
    C2N (0\%)         & 0.1660 & 0.2007 & 0.1315 & 0.1968 & 0.0581 & 0.2929 & 0.3524 & 0.2919 & 0.4517 & 0.4190 & 0.2129 & 0.2802 \\ \midrule[0.1pt]
    \textbf{Ours (0\%)} & \textbf{0.0226} & \textbf{0.1171} & \textbf{0.0218} & \textbf{0.1017} & \textbf{0.0328} & \underline{0.1176} & \textbf{0.0224} & \textbf{0.1024} & \underline{0.0284} & \textbf{0.1155} & \textbf{0.0256} & \textbf{0.1108} \\ 
    \bottomrule[0.5pt]
    \end{tabular}
    }
    \label{tab:noise_gen_per_devices}
    \vspace{-8mm}
\end{table}

\vspace{-5mm}
\subsection{sRGB Noise Generation Performance} 
\label{sec:4.2} 
\vspace{-2mm}
\subsubsection{Device-Specific Noise Quality Assessment.}
In~\ttabref{tab:noise_gen_per_devices}, we evaluate device-specific noise generation quality using KLD and AKLD. We compare \framework{} with three representative methods under different supervision settings: NAFlow~\cite{naflow} and SeNM-VAE~\cite{senm_vae}, which use paired noisy-clean supervision, and C2N~\cite{c2n}, which follows an unpaired setting. For a fair comparison, we synthesize noise using SIDD validation pairs whose metadata, such as device type and ISO, matches that of the training set.

\begin{figure*}[ht]
\begin{center}
\vspace{-7mm}
\centerline{\includegraphics[width=1.0\textwidth]{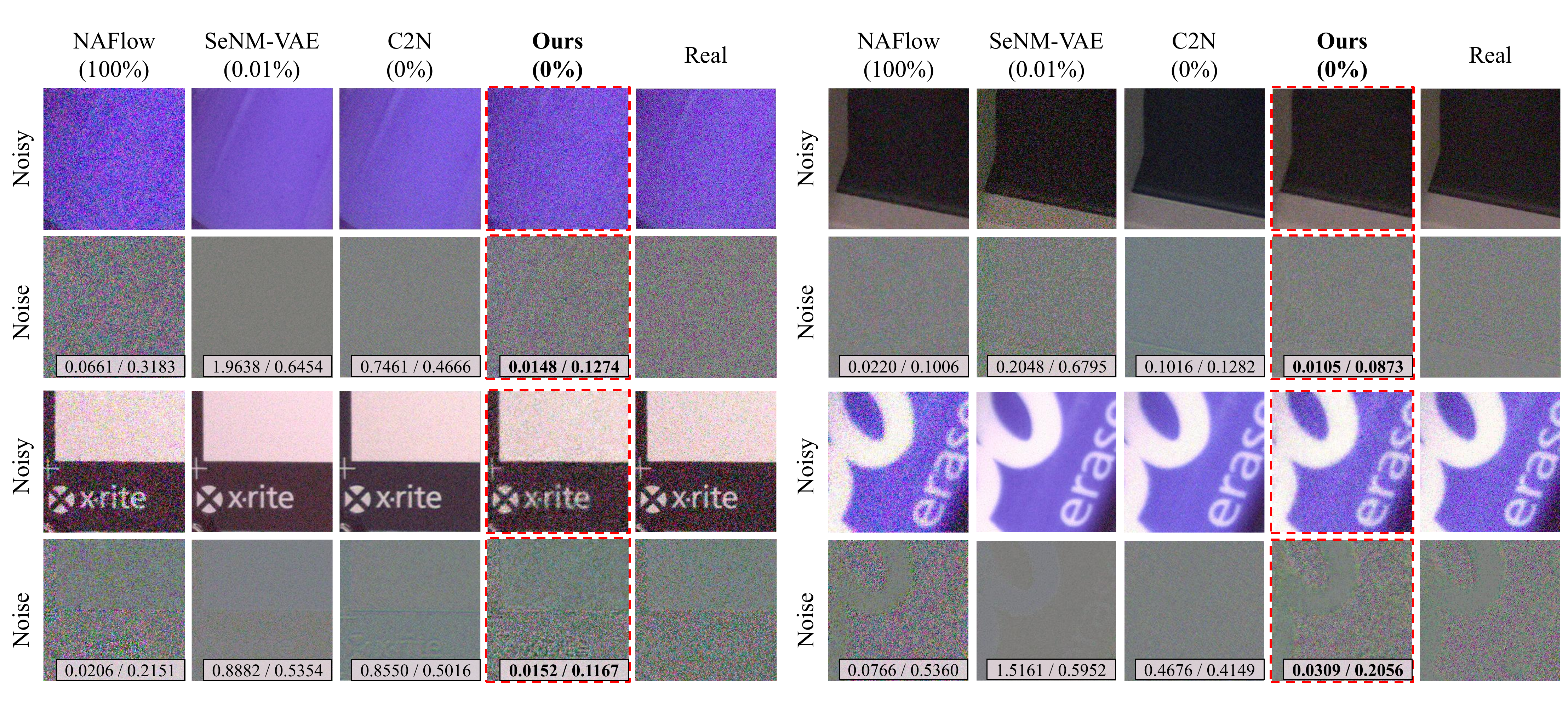}}
\vspace{-2.5mm}
\caption{Visualization of synthetic noisy images on the SIDD validation set. From left to right: NAFlow (100\%), SeNM-VAE (0.01\%), C2N (0\%), Ours (0\%, \framework{}), and real noisy images. The percentage indicates the amount of paired data used to train each noise generation model. Metrics below each image denote KLD$\downarrow$/AKLD$\downarrow$.}
\label{fig:noise_vis}
\vspace{-11mm}
\end{center}
\end{figure*}

Compared with existing methods, \framework{} achieves highly competitive performance across nearly all device types and obtains the lowest average KLD and AKLD scores. The visual results in~\ffigref{fig:noise_vis} further show that \framework{} synthesizes realistic noise patterns that closely match real-world noise in magnitude, spatial correlation, and signal dependency. These quantitative and qualitative results demonstrate that \framework{} effectively models device-specific noise distributions with high fidelity.

\vspace{-8mm}
\subsubsection{Robustness on Various Real-World Datasets.}
To further validate the robustness and generalization ability of \framework{}, we evaluate the quality of generated noise on multiple real-world datasets, including SIDD+, MAI2021, and SID. As reported in~\ttabref{tab:noise_qual_other_dataset}, \framework{} consistently achieves lower KLD and AKLD scores than NAFlow across all evaluated datasets. This is particularly notable since NAFlow requires paired noisy-clean images and camera metadata during training, whereas \framework{} does not rely on either. These results demonstrate the robustness of \framework{} across diverse camera environments and highlight its potential for augmenting real-world datasets with realistic synthetic noisy images without requiring paired ground truth or metadata.

\begin{table}[!h]
    \centering
    \vspace{-6mm}
    \caption{Quantitative results of synthetic noise on the SIDD+, MAI2021, and SID datasets (Fujifilm and Sony). All methods are trained with the SIDD training set. The results are computed with KLD$\downarrow$ and AKLD$\downarrow$. The best results are shown in \textbf{bold}.}
    \vspace{-4mm}
    \resizebox{\linewidth}{!}{ 
        \begin{tabular}{l|cccccccc|cc}
        \toprule[0.5pt]
        \multirow{2}{*}{Methods} & \multicolumn{2}{c}{SIDD+} & \multicolumn{2}{c}{MAI2021} & \multicolumn{2}{c}{Fujifilm X-T2} & \multicolumn{2}{c|}{Sony $\alpha$7S II} & \multicolumn{2}{c}{Average} \\ \cmidrule(lr){2-3} \cmidrule(lr){4-5} \cmidrule(lr){6-7} \cmidrule(lr){8-9} \cmidrule(lr){10-11}  
        & KLD$\downarrow$ & AKLD$\downarrow$ & KLD$\downarrow$ & AKLD$\downarrow$ & KLD$\downarrow$ & AKLD$\downarrow$ & KLD$\downarrow$ & AKLD$\downarrow$ & KLD$\downarrow$ & AKLD$\downarrow$ \\ \midrule[0.2pt]
        NAFlow (100\%)       & 0.0493 & 0.2914 & 0.7304 & 2.6465 & 0.3850 & 1.4640 & 0.2970 & 1.3860 & 0.3654 & 1.4470 \\
        \textbf{Ours (0\%)} & \textbf{0.0251} & \textbf{0.1451} & \textbf{0.0795} & \textbf{0.2860} & \textbf{0.1510} & \textbf{0.2410} & \textbf{0.1070} & \textbf{0.3690} & \textbf{0.0907} & \textbf{0.2603} \\ \bottomrule[0.5pt]
        \end{tabular}
    }
    \label{tab:noise_qual_other_dataset}
    \vspace{-6mm}
\end{table}

\begin{table}[!h]
        \newcommand{\tightbold}[1]{{\fontseries{b}\selectfont #1}}
        \begin{minipage}{.49\linewidth}
         \centering
         \vspace{-0.1cm}
        \caption{Quantitative denoising results on the SIDD validation dataset using different synthetic noisy data generation strategies. We report PSNR and SSIM values for two denoising backbones, AP-BSN and MM-BSN. The best and second-best results are denoted as \textbf{bold} and \underline{underline}, respectively.}
        \label{tab:sidd_denoising}
        \resizebox{0.9\textwidth}{!}{
        \begin{tabular}{lcccc}
        \toprule
        \multirow{2}{*}{Methods} & \multicolumn{2}{c}{AP-BSN} & \multicolumn{2}{c}{MM-BSN} \\
        \cmidrule(lr){2-3} \cmidrule(lr){4-5}
         & PSNR$\uparrow$ & SSIM$\uparrow$ & PSNR$\uparrow$ & SSIM$\uparrow$ \\
        \midrule
        \textit{Real Noisy}      & 36.58 & 0.8865 & \underline{37.01} & 0.8915 \\ \cmidrule(lr){1-5}
        NAFlow (100\%)          & 36.25 & 0.8912 & 36.66 & \underline{0.8941} \\ 
        SeNM-VAE (0.01\%)       & 36.58 & 0.8911 & 36.83 & 0.8868 \\ \cmidrule(lr){1-5}
        C2N (0\%)               & 34.06 & 0.8519 & 34.15 & 0.8486 \\
        Ours (0\%)              & \underline{36.86} & \underline{0.8946} & \underline{37.01} & \tightbold{0.8950} \\ \cmidrule(lr){1-5}
        SeNM-VAE-Mixed (50\%) & 36.71 & 0.8927 & 36.91 & 0.8894 \\
        \tightbold{Ours-Mixed (50\%)} & \tightbold{37.09} & \tightbold{0.8953} & \tightbold{37.09} & \tightbold{0.8950} \\
        \bottomrule
        \end{tabular}%
        }
        \end{minipage}
        \hspace{0.1cm}
        \begin{minipage}{.49\linewidth}
            \centering
            \vspace{-0.9cm}
            \caption{Quantitative denoising results on the DND benchmark dataset using different synthetic noisy data generation strategies. We report PSNR and SSIM values for two denoising backbones, AP-BSN and MM-BSN. The best and second-best results are denoted as \textbf{bold} and \underline{underline}, respectively.}
            \resizebox{1.0\textwidth}{!}{
            \begin{tabular}{lcccc}
            \toprule
            \multirow{2}{*}{Methods} & \multicolumn{2}{c}{AP-BSN} & \multicolumn{2}{c}{MM-BSN} \\
            \cmidrule(lr){2-3} \cmidrule(lr){4-5}
             & PSNR$\uparrow$ & SSIM$\uparrow$ & PSNR$\uparrow$ & SSIM$\uparrow$ \\
            \midrule
            \textit{Real Noisy}     & 38.20 & 0.939 & 38.61 & \underline{0.941} \\ \cmidrule(lr){1-5}
            \tightbold{Ours (0\%)}               & \underline{38.37} & \underline{0.940} & \underline{38.62} & \tightbold{0.942} \\
            \tightbold{Ours-Mixed (50\%)}               & \tightbold{38.55} & \tightbold{0.941} & \tightbold{38.70} & \tightbold{0.942} \\
            \bottomrule
            \end{tabular}%
            }
            \label{tab:dnd_denoising}
        \end{minipage}
        \vspace{-6mm}
\end{table}

\vspace{-6mm}
\subsection{Real-World Denoising Performance} 
\vspace{-1.5mm}
\subsubsection{Denoising Results on SIDD.}
To evaluate the effectiveness of generated noise in a downstream denoising task, we train AP-BSN and MM-BSN on synthetic noisy datasets generated from the SIDD training set. In~\ttabref{tab:sidd_denoising}, we compare self-supervised denoisers trained with \framework{}-generated images against those trained with images synthesized by competing methods, including NAFlow, SeNM-VAE, and C2N, as well as a \textit{Real Noisy} baseline. The percentages in parentheses indicate the proportion of paired clean-noisy data used to train each noise generation model.

\begin{figure*}[h]
\begin{center}
\vspace{-5mm}
\centerline{\includegraphics[width=1.0\textwidth]{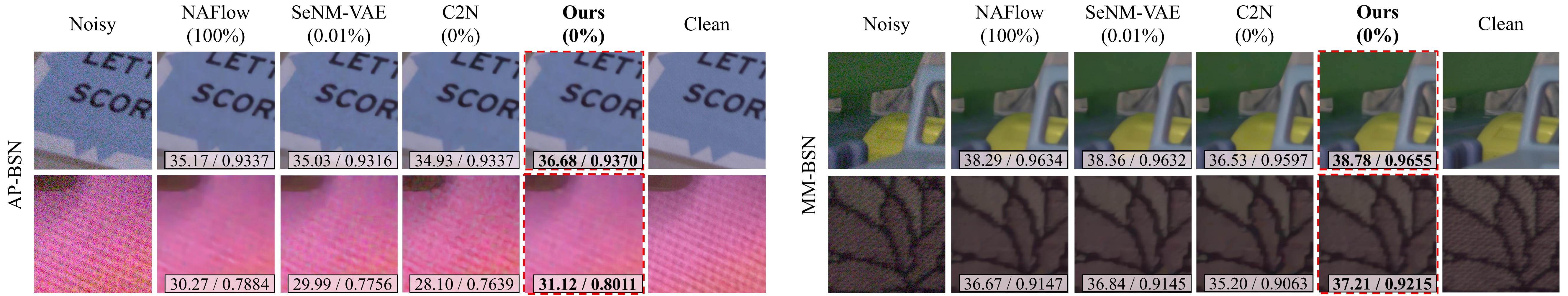}}
\vspace{-2mm}
\caption{Visual comparison. Denoising results on SIDD validation set from AP-BSN and MM-BSN trained on each method. From left to right: noisy image, NAFlow (100\%), SeNM-VAE (0.01\%), C2N (0\%), Ours (0\%, \framework{}), and clean GT. The percentage indicates the amount of paired data used to train each noise generation model. Numbers below each image denote PSNR$\uparrow$/SSIM$\uparrow$.}
\label{fig:denoise_vis}
\vspace{-14mm}
\end{center}
\end{figure*}

\framework{} consistently outperforms competing noise generators and achieves performance comparable to, or slightly better than, the \textit{Real Noisy} baseline. This indicates that \framework{} synthesizes realistic and diverse noise patterns that help reduce overfitting by exposing the denoiser to a broader range of noise realizations than the original training set. Moreover, our \textit{Mixed} strategy, which combines real and synthetic noisy images, further improves denoising performance and surpasses both the \textit{Real Noisy} baseline and other mixed-data approaches. In contrast, BSN models trained with C2N-generated data show noticeably lower performance, suggesting that its synthesized noise distribution is less aligned with the target real-world noise characteristics required for effective denoising. Qualitative results for both denoisers are shown in~\ffigref{fig:denoise_vis}.

\vspace{-5mm}
\subsubsection{Denoising Results on External Dataset.}
To evaluate the generalization of synthetic noise beyond the original training distribution, we further conduct downstream denoising experiments on the DND benchmark. Since DND does not provide clean-noisy pairs for training, \framework{} synthesizes noisy images using only the available noisy observations.
As shown in~\ttabref{tab:dnd_denoising}, AP-BSN and MM-BSN trained on \framework{}-generated data achieve performance comparable to, or slightly better than, the \textit{Real Noisy} baseline. In addition, the \textit{Mixed} strategy, which combines real and synthetic noisy images, provides further performance gains. These results indicate that \framework{} can generate effective training data for unseen real-world domains and improve denoising robustness without requiring clean ground truth.

\begin{table}[!h]
    \centering
    \vspace{-6mm}
    \begin{minipage}{.48\linewidth}
      \centering
      \caption{Quantitative results of metadata classification on SIDD validation. We evaluate the impact of Noise Encoder components on classification accuracy $\uparrow$.}
      \vspace{-2.5mm}
      \resizebox{1.0\textwidth}{!}{ 
        \begin{tabular}{l|c|cc}
          \toprule
          \multirow{2}{*}{{Methods}} & \multirow{2}{*}{{Camera Sensor (\%)}} & \multicolumn{2}{c}{{Camera Sensor + ISO}} \\ \cmidrule(lr){3-4}
          & & Top-1 (\%) & Top-3 (\%) \\ 
          \midrule[0.3pt]
          \textit{Noisy Image} & 49.98 & 41.66 & 77.14 \\ 
          \cmidrule(lr){1-4}
          w/o $\mathcal{L}_\mathrm{Norm}$    & 56.65 & 47.20 & 83.73 \\
          w/o Head  & \underline{59.70} & \underline{51.20} & \underline{85.66} \\
          \midrule[0.1pt]
          \textbf{Ours} & \textbf{76.20} & \textbf{65.22} & \textbf{94.87} \\
          \bottomrule
        \end{tabular}
      }
      \label{tab:metadata_ablation}
    \end{minipage}%
    \hfill
    \begin{minipage}{.48\linewidth}
      \centering
      \caption{Evaluation of $\mathbf{z}^\mathrm{Struct}$ purity on SIDD validation. We decode images using only $\mathbf{z}^\mathrm{Struct}$ (\ie,  $\mathbf{z}^\mathrm{Noise} = \mathbf{0}$) and compare against Clean GT.}
      \vspace{-2.5mm}
      \resizebox{0.65\textwidth}{!}{ 
        \begin{tabular}{l|cc}
          \toprule
          {Methods} & {PSNR}$\uparrow$ & {SSIM}$\uparrow$ \\ 
          \midrule[0.3pt]
          w/o $\mathcal{L}_{\mathrm{Cont}}$    & 15.49 & 0.4855 \\
          w/o Head  & 17.21 & 0.6769 \\
          w/o $\mathcal{L}_{\mathrm{TV}}$    & \underline{21.49} & \underline{0.7090} \\
          \midrule[0.1pt]
          \textbf{Ours} & \textbf{23.06} & \textbf{0.7597} \\
          \bottomrule
        \end{tabular}
      }
      \label{tab:structure_purity_ablation}
    \end{minipage}
    \vspace{-6mm}
\end{table}

\vspace{-4mm}
\subsection{Ablation Study}
\vspace{-2mm}
\subsubsection{Noise Latent Quality Evaluation.}
To examine whether $\mathbf{z}^{\mathrm{Noise}}$ captures intrinsic noise characteristics, we conduct a metadata classification task. Specifically, we train a ResNet-based classifier~\cite{resnet} to classify 16 noise profiles defined by five camera sensors and their corresponding ISO levels. As shown in~\ttabref{tab:metadata_ablation}, our full model achieves the highest accuracy, outperforming the \textit{Noisy Image} baseline. Removing either the projection head or $\mathcal{L}_{\mathrm{Norm}}$ decreases the classification accuracy, demonstrating their importance in learning a discriminative noise representation. In particular, $\mathcal{L}_{\mathrm{Norm}}$ constrains the noise latent and discourages the $\mathcal{E}_n$ from encoding redundant scene structure, guiding it to focus on stochastic noise statistics. These results indicate that $\mathcal{E}_n$ extracts noise-specific attributes that are useful for accurate real-world sRGB noise modeling.

\vspace{-4mm}
\subsubsection{Structure Latent Quality Evaluation.}
To evaluate the quality of the structure latent $\mathbf{z}^{\mathrm{Struct}}$, we perform a reconstruction test by feeding only the structure latent into the RAE decoder while setting $\mathbf{z}^{\mathrm{Noise}}$ to zero. The reconstructed image is then compared with the clean ground truth. As shown in~\ttabref{tab:structure_purity_ablation}, removing either $\mathcal{L}_{\mathrm{Cont}}$ or the projection head $\mathcal{P}$ leads to lower PSNR and SSIM, indicating weaker structure-noise disentanglement. Without these components, the model tends to rely more on $\mathbf{z}^{\mathrm{Noise}}$ for reconstruction, leaving $\mathbf{z}^{\mathrm{Struct}}$ less informative. In contrast, our full model preserves scene-specific structural information more effectively while encouraging $\mathcal{E}_n$ to focus on sensor-specific noise statistics. In addition, $\mathcal{L}_{\mathrm{TV}}$ further improves the structure representation by suppressing residual high-frequency noise.

\vspace{-5mm}
\subsubsection{Analysis on Conditioning Strategy.}
We compare our dual-conditioning strategy with a Single Encoder baseline, which uses the Noise Encoder architecture $\mathcal{E}_n$ without structural suppression through subtractive injection. As shown in~\ttabref{tab:cdit_ablation}, our method consistently achieves lower KLD and AKLD scores across all evaluated datasets. This improvement comes from the $\mathcal{E}_s$, which separates invariant scene information from noise-related features, allowing C-DiT to model the target noise distribution with less structural interference. By using $\mathbf{z}^{\mathrm{Struct}}$ as a stable structural anchor and $\mathbf{z}^{\mathrm{Noise}}$ as statistical noise guidance, \framework{} produces more realistic noisy images. The resulting improvement in noise fidelity also leads to higher downstream PSNR for AP-BSN, demonstrating that disentangled conditioning is beneficial for real-world noise modeling and restoration.

\begin{table}[h!]
\centering
\vspace{-5mm}
\caption{Ablation study on the conditioning mechanism of C-DiT on the SIDD validation, SIDD+, and MAI2021 datasets. We evaluate the efficacy of disentangled conditioning by comparing the baseline using a single latent representation against our strategy utilizing $\mathbf{z}^{\mathrm{Struct}}$ and $\mathbf{z}^{\mathrm{Noise}}$.}
\vspace{-2mm}
\newcommand{\tb}[1]{{\fontseries{b}\selectfont #1}}
\resizebox{\textwidth}{!}{
\begin{tabular}{l|cc|cc|cc}
\toprule
\multirow{2}{*}{Method} & \multicolumn{2}{c|}{SIDD} & \multicolumn{2}{c|}{SIDD+} & \multicolumn{2}{c}{MAI2021} \\
\cmidrule(lr){2-3} \cmidrule(lr){4-5} \cmidrule(lr){6-7}
 & KLD/AKLD($\downarrow$) & PSNR/SSIM($\uparrow$) & KLD/AKLD($\downarrow$) & PSNR/SSIM($\uparrow$) & KLD/AKLD($\downarrow$) & PSNR/SSIM($\uparrow$) \\
\midrule
Single Enc & 0.0275/0.1118 & 36.82/0.8942 & 0.0322/0.1624 & 35.74/0.9037 & 0.1584/0.4309 & 33.73/0.8589 \\
\tb{Ours} & \tb{0.0256}/\tb{0.1108} & \tb{36.86}/\tb{0.8946} & \tb{0.0251}/\tb{0.1451} & \tb{35.81}/\tb{0.9104} & \tb{0.0793}/\tb{0.2860} & \tb{34.40}/\tb{0.8798} \\
\bottomrule
\end{tabular}
}
\vspace{-10mm}
\label{tab:cdit_ablation}
\end{table}

\vspace{-3mm}
\section{Conclusion}
\vspace{-2mm}
We presented \framework{}, a clean-image-free and metadata-free framework for real-world sRGB noise generation. Unlike prior approaches that rely on paired clean-noisy images or camera metadata, \framework{} learns from only two noisy observations of the same scene during training and requires only a single noisy image at inference. By disentangling scene structure and noise characteristics with a Reconstruction Autoencoder (RAE) and modeling the latent noise distribution using a one-step Conditional DiT (C-DiT) trained with consistency objectives, \framework{} synthesizes realistic, signal-dependent noise while preserving the underlying scene content. Extensive experiments demonstrate that \framework{} effectively captures device-specific noise distributions and generalizes across diverse real-world sensor domains. Moreover, self-supervised denoisers trained with \framework{}-synthesized data achieve competitive performance compared to those trained on real noisy images, confirming the practical value of scalable noise generation without clean references or metadata. These results suggest that two-noisy-observation-based noise modeling is a promising direction for building robust real-world image restoration systems.

\vspace{-3mm}
\section*{Acknowledgments}
\vspace{-2mm}
This research was supported by Culture, Sports and Tourism R\&D Program through the Korea Creative Content Agency grant funded by the Ministry of Culture, Sports and Tourism in 2026(Project Name: Development of AI-Based Animation Production Technology to Ensure Character and Scene Consistency and Continuity for Enhanced Efficiency and Quality, Project Number: RS-2026-25525207, Contribution Rate: 30\%). 
This work was also supported by Institute of Information \& communications Technology Planning \& Evaluation (IITP) grant funded by the Korea government (MSIT) (No.2022-0-00156, Fundamental research on continual meta-learning for quality enhancement of casual videos and their 3D metaverse transformation), 
IITP grant funded by the Korea government (MSIT) (No.RS-2020-II201373, Artificial Intelligence Graduate School Program (Hanyang University)).

%
%
\clearpage
\bibliographystyle{splncs04}
\bibliography{main}

\renewcommand{\thesection}{S\arabic{section}}
\renewcommand{\thesubsection}{S\arabic{section}.\arabic{subsection}}

\renewcommand{\thefigure}{S\arabic{figure}}
\renewcommand{\thetable}{S\arabic{table}}
\setcounter{figure}{0}
\setcounter{table}{0}
\clearpage
\setcounter{page}{1}

\begin{center}
    {\Large\bfseries YeTI: You Only Need Two Noisy Images for Real-World sRGB Noise Generation \par}
    \vspace{4mm}
    {\Large\bfseries Supplementary Material \par}
\end{center}

\begin{figure*}[!ht]
\begin{center}
\centerline{\includegraphics[width=0.9\textwidth]{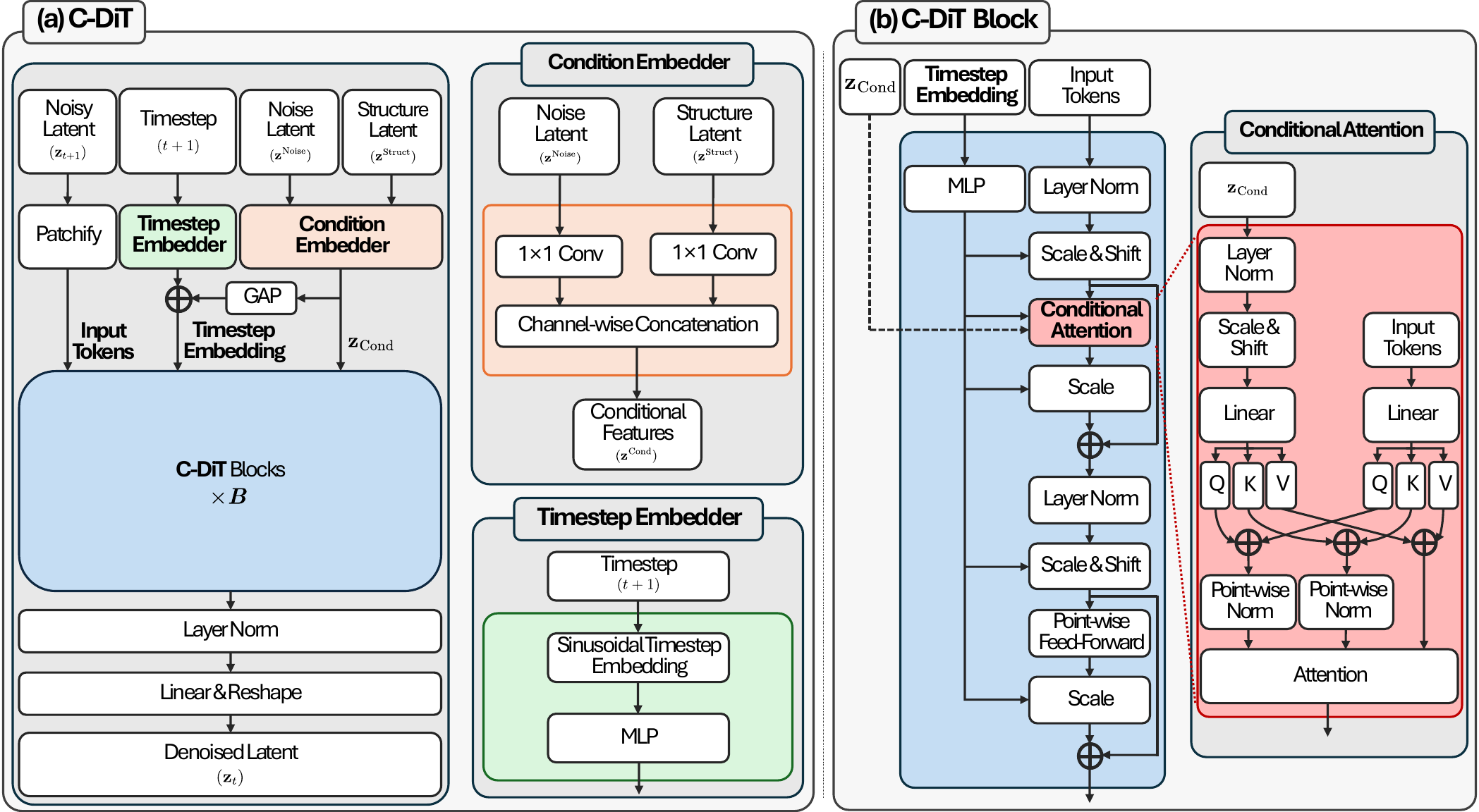}}
\caption{(a) C-DiT overview. (b) C-DiT block.}
\vspace{-18mm}
\label{fig:cdit_supple}
\end{center}
\end{figure*}

\section{C-DiT}
\label{suppl_sec_cdit}

As illustrated in \ffigref{fig:cdit_supple} (a), the overall Conditional Diffusion Transformer (C-DiT) is mainly composed of C-DiT Blocks and takes Input Tokens, Timestep Embedding, and conditioning feature $\mathbf{z}^{\mathrm{Cond}}$ as inputs. 
To be specific, Input tokens are obtained by patchifying the input latent $\mathbf{z}_{t+1}$ at time step $(t+1)$.
In addition, the noise latent $\mathbf{z}^{\mathrm{Noise}}$ and structure latent $\mathbf{z}^{\mathrm{Struct}}$
from the Noise Encoder and the Structure Encoder undergo $1{\times}1$ convolutions and channel-wise concatenation to produce the conditioning feature $\mathbf{z}^{\mathrm{Cond}}$. 
Furthermore, to secure global context, $\mathbf{z}^{\mathrm{Cond}}$ is compressed using global average pooling (GAP) and then integrated with the sinusoidal timestep embedding, yielding the Timestep Embedding.  
After $B$ C-DiT Blocks, the output tokens are further processed with Layer Norm and a Linear \& Reshape operation to output the denoised latent ($\mathbf{z}_{t}$) in the timestep $(t)$.

To preserve fine-grained spatial information, the detailed C-DiT block depicted in \ffigref{fig:cdit_supple} (b) incorporates the Conditional Attention mechanism. The timestep embedding is processed by a multi-layer perceptron (MLP) to generate Scale \& Shift parameters for Adaptive Layer Normalization (AdaLN)~\cite{dit}. 
The Input Tokens are modulated by these parameters and passed to the Conditional Attention module. 
Within this module, both the modulated input tokens and the $\mathbf{z}^{\mathrm{Cond}}$ are independently projected into query ($\mathbf{Q}$), key ($\mathbf{K}$), and value ($\mathbf{V}$) representations. Specifically, $\mathbf{z}^{\mathrm{Cond}}$ undergoes Layer Normalization and a Scale \& Shift operation before its linear projection. The corresponding $\mathbf{Q}$, $\mathbf{K}$, and $\mathbf{V}$ features from both branches are then fused respectively via element-wise addition. Following this pairwise integration, the combined $\mathbf{Q}$ and $\mathbf{K}$ undergo point-wise normalization prior to the standard attention computation. Finally, the output features from the attention module are further refined through a point-wise feed-forward network to synthesize the precise noise distribution.

\vspace{-2mm}
\section{Training Details}
\label{suppl_training}

\subsection{RAE Training}
\noindent\textbf{Optimization and Hyperparameters.} 
The balancing weights for the RAE objective defined in Eq. (9) of the main manuscript are set as follows: $\lambda_{\mathrm{Cont}} = 1.0$, $\lambda_{\mathrm{TV}} = 1\text{e-}6$, and $\lambda_{\mathrm{Norm}} = 1\text{e-}6$. For the InfoNCE contrastive objective~\cite{infonce}, a temperature scaling parameter of $\tau = 0.7$ is applied on the normalized features. As detailed in Sec. 4.1 of the main manuscript, the model is trained utilizing the Adam optimizer~\cite{adam}, and the specific momentum parameters are set to $\beta_1 = 0.5$ and $\beta_2 = 0.9$, with an epsilon value of 1\text{e-}8 to ensure numerical stability.

\noindent\textbf{EMA Teacher.} 
For stable training with contrastive learning, we update the teacher structure encoder using an exponential moving average (EMA)~\cite{moco} of the corresponding student encoder.
The parameters of this teacher model are updated at each training step with a momentum decay rate of $\gamma = 0.999$.

\noindent\textbf{Gradient Detachment for Disentanglement.} 
To disentangle the noise and structure, and enforce latent disentanglement, the structural prior extracted by the Structure Encoder ($\mathcal{E}_s$) is detached from the computational graph before being injected into the Noise Encoder ($\mathcal{E}_n$). Without this stop-gradient operation, the gradients derived from reconstruction loss would backpropagate through the injection pathway into $\mathcal{E}_s$, inadvertently forcing the student encoder to capture noise-specific degradation signals. 

\vspace{-2mm}
\subsection{C-DiT Training}

\noindent\textbf{Optimization and iCT Framework.} 
C-DiT is optimized using the RAdam optimizer~\cite{radam} with momentum parameters set to $\beta_1 = 0.9$ and $\beta_2 = 0.999$, along with an epsilon value of 1e-8. 
To learn a one-step generative mapping, we adopt the Improved Consistency Training (iCT) framework~\cite{ict} and minimize the pseudo-Huber loss~\cite{huber_loss}. 
Although we strictly follow the default continuous noise schedule and lognormal timestep sampling proposed in iCT, we specifically modify the discretization curriculum to dynamically increase from 10 to 40 steps. 
This reduced maximum step count is explicitly tailored to our highly compressed $32{\times}32$ latent space; since the generative trajectory in this compact space is significantly less complex than in the high-dimensional pixel space, fewer discretization steps are required for stable convergence.

\noindent\textbf{Latent Preconditioning.} 
As a standard pre-processing step to ensure numerical stability during the diffusion process~\cite{edm}, the raw latent representations extracted from RAE undergo channel-wise preconditioning. The empirical mean and standard deviation of each channel are explicitly shifted and scaled to match a target data distribution with zero-mean and a standard deviation of 0.5. This ensures that the linear scaling is applied uniformly to all samples to maintain the inherent contrast and discriminative signals of the raw latents while facilitating stable generative modeling.

\vspace{-2mm}
\section{Architecture Details}
\label{suppl_architecture}

\noindent\textbf{Overall Architecture of RAE.} 
The Structure Encoder ($\mathcal{E}_s$), Noise Encoder ($\mathcal{E}_n$), and Decoder ($\mathcal{D}$) share a symmetric multi-scale design. The base channel dimension is set to $C = 48$ with $3{\times}3$ convolutional kernels. The feature extraction spans four spatial scales, employing a single Residual Dense Channel Attention (RDCA) block , as depicted in Fig. 3(b) of the main manuscript, at each level, which ultimately compresses the input into a latent bottleneck of $32{\times}32$ spatial resolution with 192 channels.

\noindent\textbf{Contrastive Projection Head.} 
The projection head $\mathcal{P}(\cdot)$ is implemented as an Attentional Pooling~\cite{clip} module. It aggregates the flattened $32{\times}32$ spatial features into a compact 192-dimensional embedding using 8 attention heads.

\noindent\textbf{Network Backbone of C-DiT.} 
The C-DiT backbone employs a Transformer architecture~\cite{attention, mmdit} comprising 8 layers, a hidden dimension of 512, and 8 attention heads with an MLP expansion ratio of 4.0. Operating on the heavily compressed $32{\times}32$ latent space with 192 channels, the input representation is tokenized using a patch size of $1{\times}1$ pixel without further spatial downsampling.

\vspace{-2mm}
\section{Model Size and Inference Speed}
\label{suppl_model_size}

\vspace{-10mm}
\begin{table}[h!]
    \centering
    \caption{Comparison of model complexity, inference speed, and generation metrics on the SIDD validation dataset. MACs and Inference Time are measured for a single $256{\times}256$ image generation on an RTX A5000 GPU.}
    \vspace{-2mm}
    \resizebox{0.8\textwidth}{!}{
        \begin{tabular}{l|ccc|cc}
        \toprule
        Methods & Params (M) & MACs (G) & Inference Time (ms) & KLD$\downarrow$ & AKLD$\downarrow$ \\
        \midrule
        NAFlow & 1.115 & 4.750 & 72.94 & 0.0305 & 0.1306 \\
        SeNM-VAE (0.01\%) & 19.96 & 154.34 & 52.90 & 0.0672 & 0.1295 \\
        C2N & 2.154 & 70.53 & 18.65 & 0.2129 & 0.2802 \\
        \midrule
        Ours (Light) & 23.89 & 13.97 & 16.64 & 0.0309 & 0.1240 \\
        \textbf{Ours} & 64.77 & 44.52 & 20.79 & \textbf{0.0256} & \textbf{0.1108} \\
        \bottomrule
        \end{tabular}
    }
    \vspace{-5mm}
    \label{tab:model_complexity}
\end{table}

We evaluate the computational complexity, inference speed, and generation quality of \framework{} compared to existing sRGB noise generation models.
\framework{} processes inputs through a RAE and a C-DiT. 
By operating within a highly compact latent space rather than the high-dimensional pixel space, our framework significantly reduces the computational bottleneck of the diffusion process, thereby improving the overall inference efficiency. 
\ttabref{tab:model_complexity} reports the number of parameters, MACs, inference time and generation metrics (KLD/AKLD). In particular, the inference time is measured for generating a single $256{\times}256$ image on an NVIDIA RTX A5000 GPU.
The results indicate that despite having a larger parameter size, \framework{} achieves inference speeds and computational costs (MACs) highly competitive with existing models like C2N, SeNM-VAE, and NAFlow, owing to its efficient latent generative framework.
Additionally, we introduce a lightweight variant, \framework{} (Light). It achieves the fastest inference speed among all evaluated methods, yet still surpasses the paired supervision baseline, SeNM-VAE, in generation fidelity.

\vspace{-2mm}
\section{Additional Denoising Performance}
\label{suppl_additional_denoising}

\vspace{-10mm}
\begin{table}[h!]
    \centering
    \newcommand{\tightbold}[1]{{\fontseries{b}\selectfont #1}}
    \caption{BSN denoising evaluation on the SIDD validation set across three random seeds (mean $\pm$ std). The best mean and standard deviation results are highlighted in \textbf{bold} individually.}
    \vspace{-3mm}
    \resizebox{0.85\textwidth}{!}{
        \begin{tabular}{l|cc|cc}
        \toprule
        \multirow{2}{*}{Methods} & \multicolumn{2}{c|}{AP-BSN} & \multicolumn{2}{c}{MM-BSN} \\
        & PSNR ($\uparrow$) & SSIM ($\uparrow$) & PSNR ($\uparrow$) & SSIM ($\uparrow$) \\
        \midrule
        Real-Noisy (GT) & 36.75 $\pm$ 0.2194 & 0.8899 $\pm$ 0.0030 & 37.02 $\pm$ \tightbold{0.0148} & 0.8924 $\pm$ 0.0007 \\
        \tightbold{Ours} & \tightbold{36.96} $\pm$ \tightbold{0.1102} & \tightbold{0.8934} $\pm$ \tightbold{0.1102} & \tightbold{37.03} $\pm$ 0.0153 & \tightbold{0.8925} $\pm$ \tightbold{0.0006} \\
        \bottomrule
        \end{tabular}
    }
    \vspace{-5mm}
    \label{tab:seed_variance}
\end{table}

To thoroughly assess the stability and robustness of the generated noise for downstream tasks, we evaluate the denoising performance across three different random seeds. We train self-supervised denoisers, AP-BSN and MM-BSN, using the synthetic noise generated by \framework{} and compare them against the \textit{Real-Noisy (GT)} baseline. As reported in \ttabref{tab:seed_variance}, the models trained on our synthetic noise consistently achieve mean PSNR and SSIM scores that are highly competitive with, or slightly improving upon, the real noisy data. Furthermore, \framework{} exhibits standard deviations that are comparable to the real noisy baseline across independent runs, indicating that our generated noise distributions provide stable and reliable training signals regardless of initialization.

Additionally, to demonstrate the utility of our synthetic data beyond self-supervised settings, we evaluate its effectiveness for fully supervised denoising. We train a widely adopted supervised model, DnCNN~\cite{dncnn}, using clean images paired with noise synthesized by \framework{} and the baseline, SeNM-VAE. As reported in \ttabref{tab:dncnn_results}, the model trained on our synthetic noise yields a substantial performance margin over the variant trained with SeNM-VAE. This indicates that \framework{} synthesizes effective and realistic noise patterns that serve as robust supervisory signals for supervised architectures, despite requiring no paired real-world data during training.

\begin{table}[h!]
    \vspace{-5mm}
    \centering
    \newcommand{\tightbold}[1]{{\fontseries{b}\selectfont #1}}
    \begin{minipage}[t]{.48\linewidth}
        \centering
        \caption{Evaluation of supervised denoising performance using DnCNN on the SIDD validation set.}
        \vspace{-2mm}
        \resizebox{0.85\textwidth}{!}{
            \begin{tabular}{l|cc}
            \toprule
            Methods & PSNR$\uparrow$ & SSIM$\uparrow$ \\
            \midrule
            SeNM-VAE (0.01\%) & 35.82 & 0.8715 \\
            \tightbold{Ours} & \tightbold{36.32} & \tightbold{0.8821} \\
            \bottomrule
            \end{tabular}
        }
        \label{tab:dncnn_results}
    \end{minipage}\hfill
    \begin{minipage}[t]{.48\linewidth}
        \centering
        \caption{Evaluation of $\mathbf{z}^\mathrm{Struct}$ purity on SIDD val. Decoded using only $\mathbf{z}^\mathrm{Struct}$ (\ie $\mathbf{z}^\mathrm{Noise} = \mathbf{0}$).}
        \vspace{-2mm}
        \resizebox{0.44\textwidth}{!}{ 
            \begin{tabular}{l|cc}
            \toprule
            Methods & PSNR$\uparrow$ & SSIM$\uparrow$ \\ 
            \midrule
            Add & 12.70 & 0.2576 \\
            Concat & 14.42 & 0.6315 \\
            \midrule
            \tightbold{Ours} & \tightbold{23.06} & \tightbold{0.7597} \\
            \bottomrule
            \end{tabular}
        }
        \label{tab:suppl_structure_injection}
    \end{minipage}
    \vspace{-10mm}
\end{table}

\vspace{-2mm}
\section{Ablation on Injection Module}
\label{suppl_ablation}

\vspace{-2mm}
\begin{figure*}[h!]
\begin{center}
\vspace{-5mm}
\centerline{\includegraphics[width=1.0\textwidth]{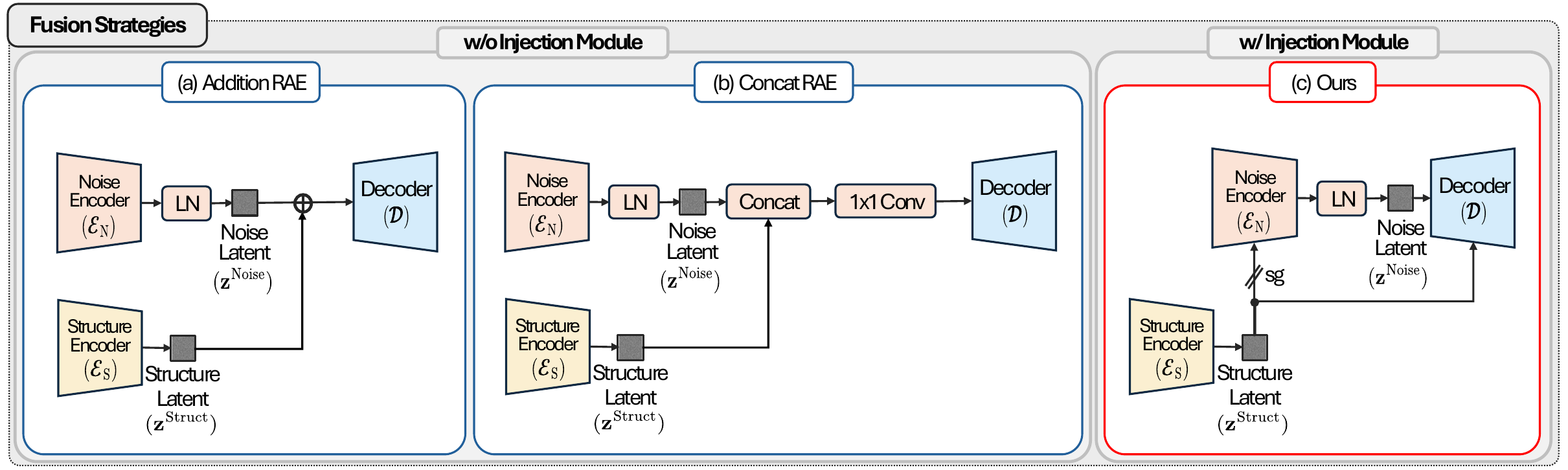}}
\caption{Architectural variants for the injection ablation study. To explicitly evaluate the impact of the injection module on latent disentanglement, the diagrams illustrate (a) the Addition baseline utilizing element-wise addition, (b) the Concatenation baseline employing channel concatenation, and (c) the proposed asymmetric injection mechanism.}
\vspace{-5mm}
\label{fig:add_concat_supple}
\end{center}
\end{figure*}

We evaluate the impact of the proposed injection module on latent disentanglement within the RAE. As depicted in \ffigref{fig:add_concat_supple}, we compare our module against two baselines: element-wise addition and channel concatenation. Disentanglement quality is quantified by structural fidelity, measured by clean PSNR and SSIM when decoding solely from the structural latent ($\mathbf{z}^{\mathrm{Struct}}$). 

\vspace{-8mm}
\begin{figure*}[h!]
\begin{center}
\centerline{\includegraphics[width=0.7\textwidth]{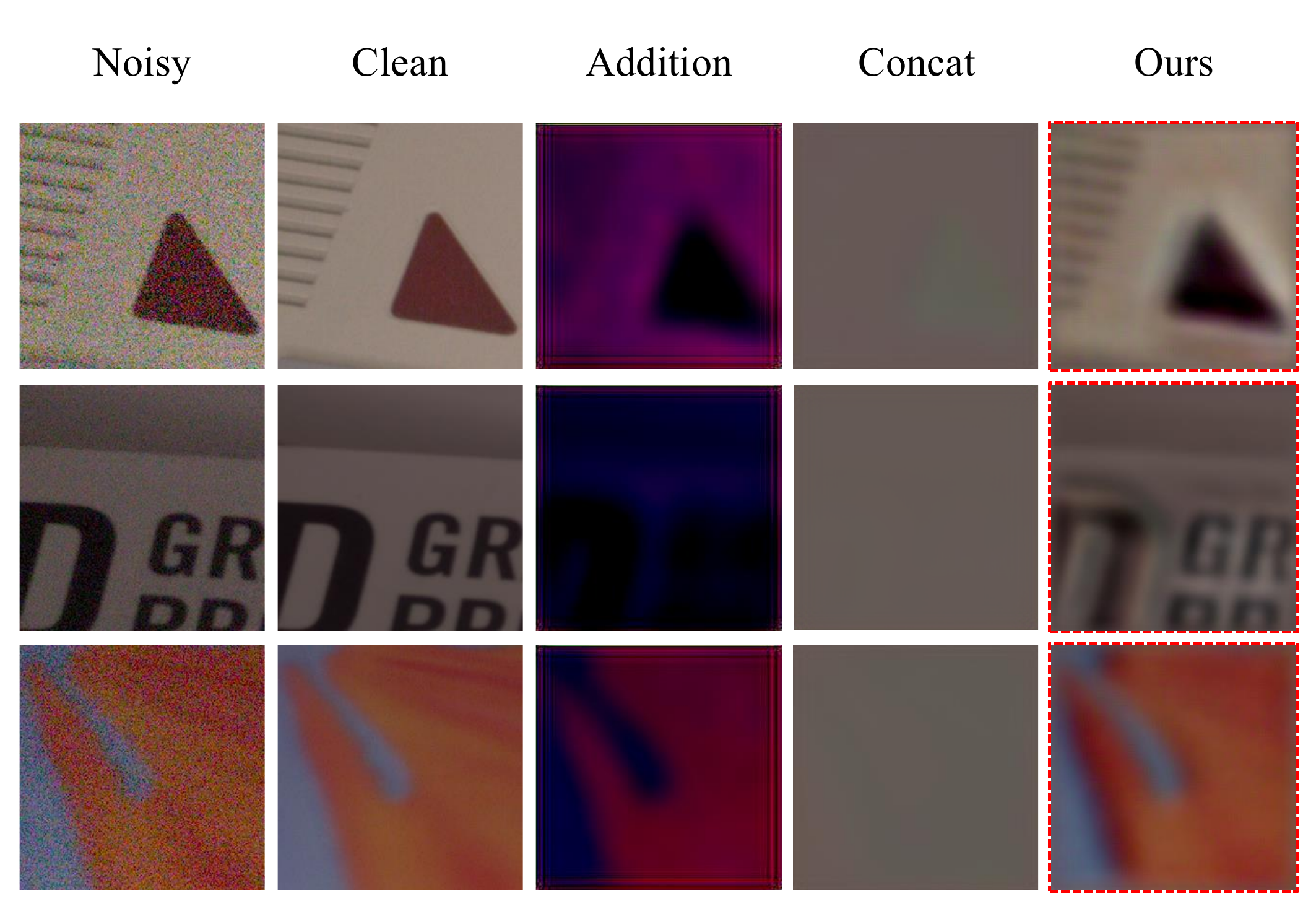}}
\caption{Qualitative evaluation of structural fidelity across fusion strategies. The sequence from left to right displays the noisy input, the clean ground truth, and the decoding results utilizing solely the structural latent $\mathbf{z}^\mathrm{Struct}$ by setting $\mathbf{z}^\mathrm{Noise}$ to zero for the addition baseline, the concatenation baseline, and the proposed injection module.}
\vspace{-10mm}
\label{fig:zero_latent_supple}
\end{center}
\end{figure*}

As quantitatively reported in \ttabref{tab:suppl_structure_injection} and visually demonstrated in \ffigref{fig:zero_latent_supple}, the addition and concatenation methods fail to preserve the structural prior. The addition strategy induces severe representation collapse and color distortion. Concatenation suffers from a severe spatial information bottleneck during channel compression, yielding a featureless flat image. Conversely, our injection decouples the representations via explicit subtraction in the encoder, avoiding structural collapse or color fidelity loss, and achieves the highest clean PSNR of 23.06 dB.

\section{Ablation on Hyperparameters}
\label{suppl_loss_ablation}

To find the optimal balance between noise disentanglement and structural preservation within the RAE, we conduct an ablation study on the loss balancing weights, specifically $\lambda_{\mathrm{TV}}$ and $\lambda_{\mathrm{Norm}}$.
As reported in \ttabref{tab:suppl_loss_weights}, setting both weights to 1e-6 yields the optimal performance. 
Applying larger weights over-regularizes the structural and sensor-specific features, which impedes the extraction of pure sensor statistics and consequently reduces metadata classification accuracy. Conversely, smaller weights insufficiently guide the latent disentanglement, leading to an ineffective information bottleneck that degrades structural purity during clean-signal reconstruction. 
Therefore, we establish 1e-6 as the default setting, achieving the best trade-off between accurately capturing device-specific noise attributes and preserving invariant scene structures.

\begin{table}[h!]
    \centering
    \newcommand{\tightbold}[1]{{\fontseries{b}\selectfont #1}}
    \caption{Ablation of loss weights ($\lambda_{\mathrm{TV}}$ and $\lambda_{\mathrm{Norm}}$) on the SIDD validation set. We report metadata classification accuracy for noise characteristics and PSNR/SSIM for structure purity. The best and second-best results are highlighted in \tightbold{bold} and \underline{underline}, respectively.}
    \vspace{2mm}
    \setlength{\tabcolsep}{8pt} 
    \renewcommand{\arraystretch}{1.1}
    \resizebox{0.85\textwidth}{!}{
        \begin{tabular}{cc|ccc|cc}
        \toprule
        $\lambda_{\mathrm{TV}}$ & $\lambda_{\mathrm{Norm}}$ & Camera & \multicolumn{2}{c|}{Camera Sensor + ISO} & \multicolumn{2}{c}{Structure Purity} \\
        \cmidrule(lr){4-5} \cmidrule(lr){6-7}
        Weight & Weight & Sensor (\%) & Top-1 (\%) & Top-3 (\%) & PSNR ($\uparrow$) & SSIM ($\uparrow$) \\
        \midrule
        1e-7 & 1e-7 & 61.62 & 52.56 & 86.94 & 21.74 & \underline{0.7300} \\
        \tightbold{1e-6} & \tightbold{1e-6} & \tightbold{76.20} & \tightbold{65.22} & \tightbold{94.87} & \underline{23.06} & \tightbold{0.7597} \\
        1e-5 & 1e-5 & 50.08 & 42.71 & 80.29 & \tightbold{23.15} & 0.7250 \\
        \bottomrule
        \end{tabular}
    }
    \label{tab:suppl_loss_weights}
\end{table}

\section{Robustness to Spatial Misalignment}
\label{suppl_misalignment}

Real-world burst capture may occasionally introduce slight spatial misalignments between frames due to camera motion. To evaluate robustness against such artifacts, we simulate burst misalignment by shifting the paired noisy input frames by $N$ pixels ($N \in \{2, 4\}$) during training. 

As shown in \ttabref{tab:suppl_misalignment}, \framework{} with $N=2$ remains highly consistent with the perfectly aligned setting ($N=0$). Furthermore, even under a more severe shift ($N=4$), our method achieves downstream denoising performance highly competitive with the perfectly aligned baseline, SeNM-VAE ($N=0$). This indicates that \framework{} is inherently robust to small spatial misalignments, effectively resolving them through its robust latent structure and noise decomposition.

\begin{table}[h!]
    \centering
    \newcommand{\tightbold}[1]{{\fontseries{b}\selectfont #1}}
    \caption{Effect of $N$-pixel misalignment on the SIDD validation set. We evaluate the robustness of noise generation (KLD/AKLD) and downstream denoising performance (PSNR/SSIM) using AP-BSN.}
    \vspace{2mm}
    \setlength{\tabcolsep}{8pt} 
    \renewcommand{\arraystretch}{1.1}
    \resizebox{0.85\textwidth}{!}{
        \begin{tabular}{l|c|ccc}
        \toprule
        \multirow{2}{*}{Metric} & SeNM-VAE (0.01\%) & \multicolumn{3}{c}{\tightbold{Ours}} \\
        & ($N=0$) & ($N=4$) & ($N=2$) & ($N=0$) \\
        \midrule
        KLD / AKLD ($\downarrow$) & 0.0672 / 0.1295 & 0.0416 / 0.1549 & 0.0413 / 0.1210 & \tightbold{0.0256} / \tightbold{0.1108} \\
        PSNR / SSIM ($\uparrow$) & 36.58 / 0.8911 & 36.66 / 0.8918 & 36.79 / 0.8924 & \tightbold{36.86} / \tightbold{0.8946} \\
        \bottomrule
        \end{tabular}
    }
    \label{tab:suppl_misalignment}
\end{table}

\section{Analysis of Signal-Dependent Noise}
\label{suppl_signal_dependent}

Real-world sRGB noise exhibits complex signal-dependent characteristics, where the noise variance varies non-linearly with the underlying image intensity. To verify whether \framework{} captures this physical property, we plot the relationship between noise variance and the mean intensity of the clean images across five different camera devices from the SIDD validation set.

\begin{figure*}[h!]
\begin{center}
\vspace{-2mm}
\centerline{\includegraphics[width=1.0\textwidth]{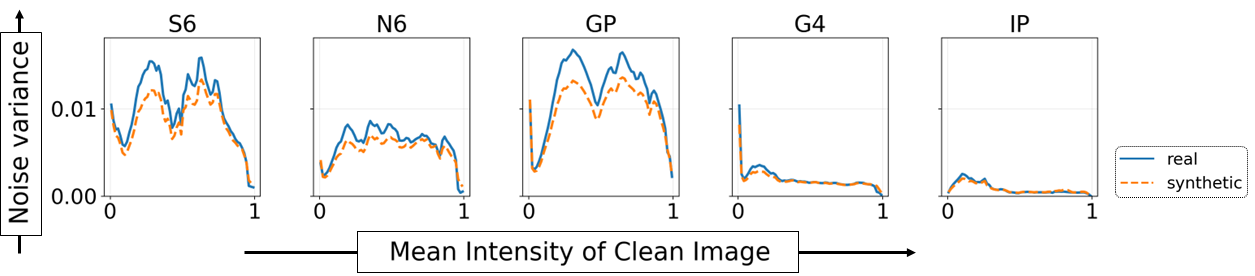}}
\caption{Analysis of signal-dependent noise characteristics. The plots illustrate the relationship between noise variance and mean image intensity across five different camera devices on the SIDD validation set. The synthetic noise generated by \framework{} follows the real noise variance-intensity trend.}
\vspace{-10mm}
\label{fig:variance_intensity}
\end{center}
\end{figure*}

As illustrated in \ffigref{fig:variance_intensity}, the variance profile of our synthetic noise (dashed orange line) follows the real SIDD variance-intensity trend (solid blue line) across the intensity spectrum. This behavior is consistently observed across all five distinct camera sensors (S6, N6, GP, G4, and IP), despite their varied device-specific noise curves. These results suggest that \framework{} is capable of capturing both the global metadata-dependent characteristics and the fine-grained signal-dependent noise distributions present in real-world scenarios.

\vspace{-2mm}
\section{Visualizing the Disentangled Latents}

To explicitly verify whether RAE successfully decouples the noisy input image into independent noise and structural components, a qualitative visualization experiment is conducted on the intermediate latent spaces.
To project the high-dimensional latent tensors into the two-dimensional spatial domain for visual inspection, the extracted features $\mathbf{z}^{\mathrm{Noise}}$ and $\mathbf{z}^{\mathrm{Struct}}$ are aggregated via channel-wise average pooling. The aggregated spatial maps are subsequently resized using bicubic interpolation and normalized through instance-level min-max normalization.

\begin{figure*}[h!]
\begin{center}
\centerline{\includegraphics[width=0.7\textwidth]{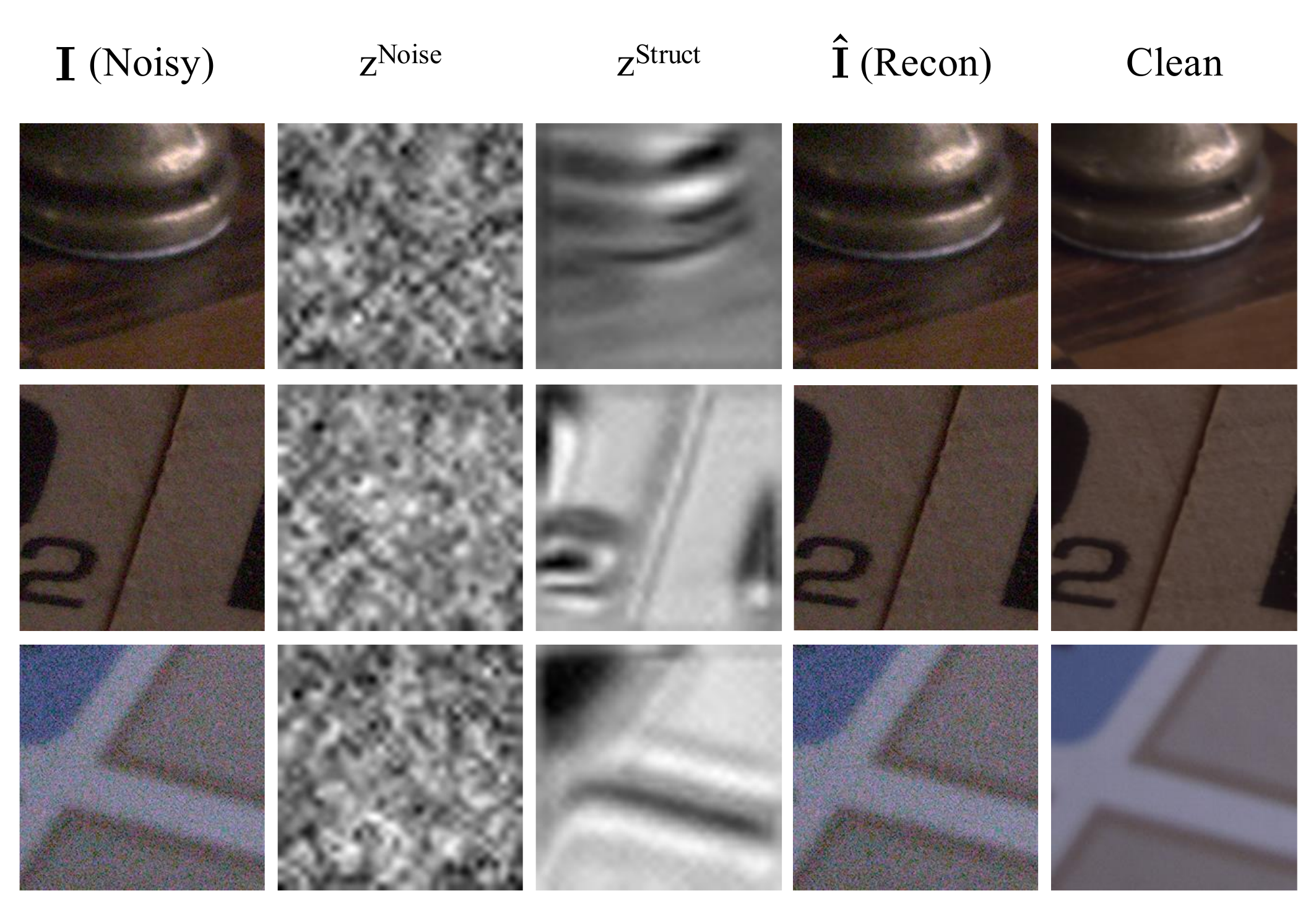}}
\caption{Qualitative visualization of the disentangled latent representations. From left to right: noisy input, noise latent $\mathbf{z}^{\mathrm{Noise}}$, structural latent $\mathbf{z}^{\mathrm{Struct}}$, reconstructed output, and the GT clean image.}
\vspace{-12mm}
\label{fig:rae_latent_supple}
\end{center}
\end{figure*}

As illustrated in \ffigref{fig:rae_latent_supple}, the visualization clearly demonstrates how the input is decomposed into two distinct latents. The structural latent $\mathbf{z}^{\mathrm{Struct}}$ faithfully retains the prominent structural patterns and object shapes of the original scene. In contrast, the noise latent $\mathbf{z}^{\mathrm{Noise}}$ predominantly captures spatially unstructured noise patterns, effectively suppressing the explicit structural features of the underlying scene. Integrating these decoupled representations yields a faithful reconstruction of the original noisy input. 
These visual results directly substantiate the robust disentanglement capability of our method.

\vspace{-5mm}

\section{More Qualitative Results}
\label{suppl_qualitative}

\subsection{Qualitative Results of Noise Generation}
\begin{figure*}[t]
\begin{center}
\centerline{\includegraphics[width=1.0\textwidth]{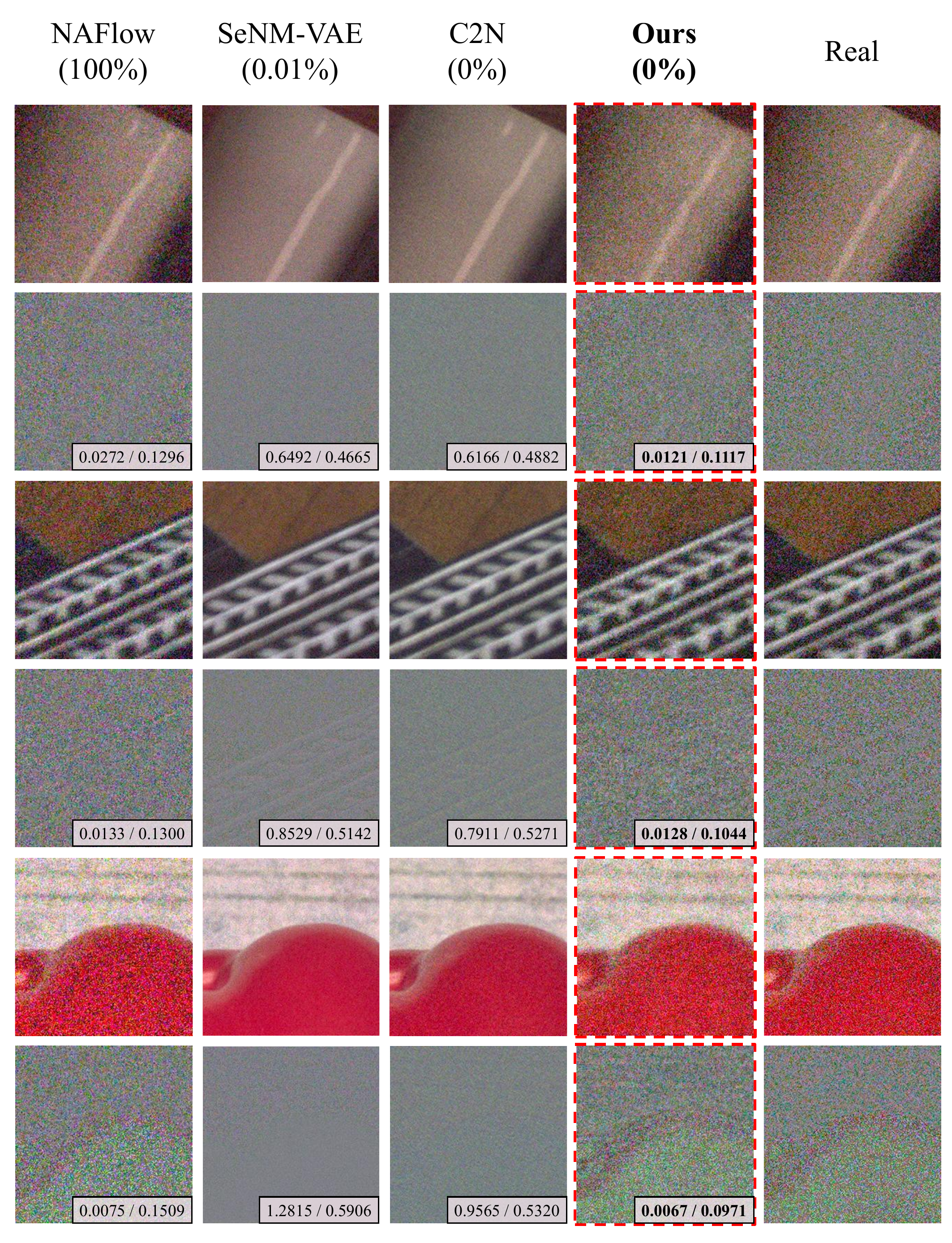}}
\caption{Visualization of synthetic noisy images on the SIDD validation set. From left to right: NAFlow, SeNM-VAE, C2N, Ours (\framework{}),
and real noisy images. The percentage indicates the amount of paired data used to train each noise generation model. Numbers below each image denote KLD / AKLD.}
\vspace{-12mm}
\label{fig:noise_visual_supple}
\end{center}
\end{figure*}

In \ffigref{fig:noise_visual_supple}, we provide additional visualizations of generated noise, comparing our method (\framework{}) with other approaches: NAFlow~\cite{naflow}, SeNM-VAE~\cite{senm_vae}, and C2N~\cite{c2n}.
As shown in the visual comparisons, while NAFlow succeeds in generating noise with a perceptible magnitude, its synthesized patterns still exhibit noticeable visual discrepancies and fail to fully capture the authentic spatial structures of real-world noise. Furthermore, other models such as SeNM-VAE and C2N tend to synthesize weak and insufficient noise that fails to represent the true intensity and complex signal-dependent characteristics of real sRGB noise. In contrast, \framework{} successfully synthesizes highly realistic noise that accurately reflects both the spatial correlation and magnitude of real-world sensor noise, strictly preserving the underlying color and scene structures without requiring any paired clean images.

\vspace{-2mm}
\subsection{Qualitative Results of Denoising Performance}

\begin{figure*}[t]
\begin{center}
\centerline{\includegraphics[width=0.9\textwidth]{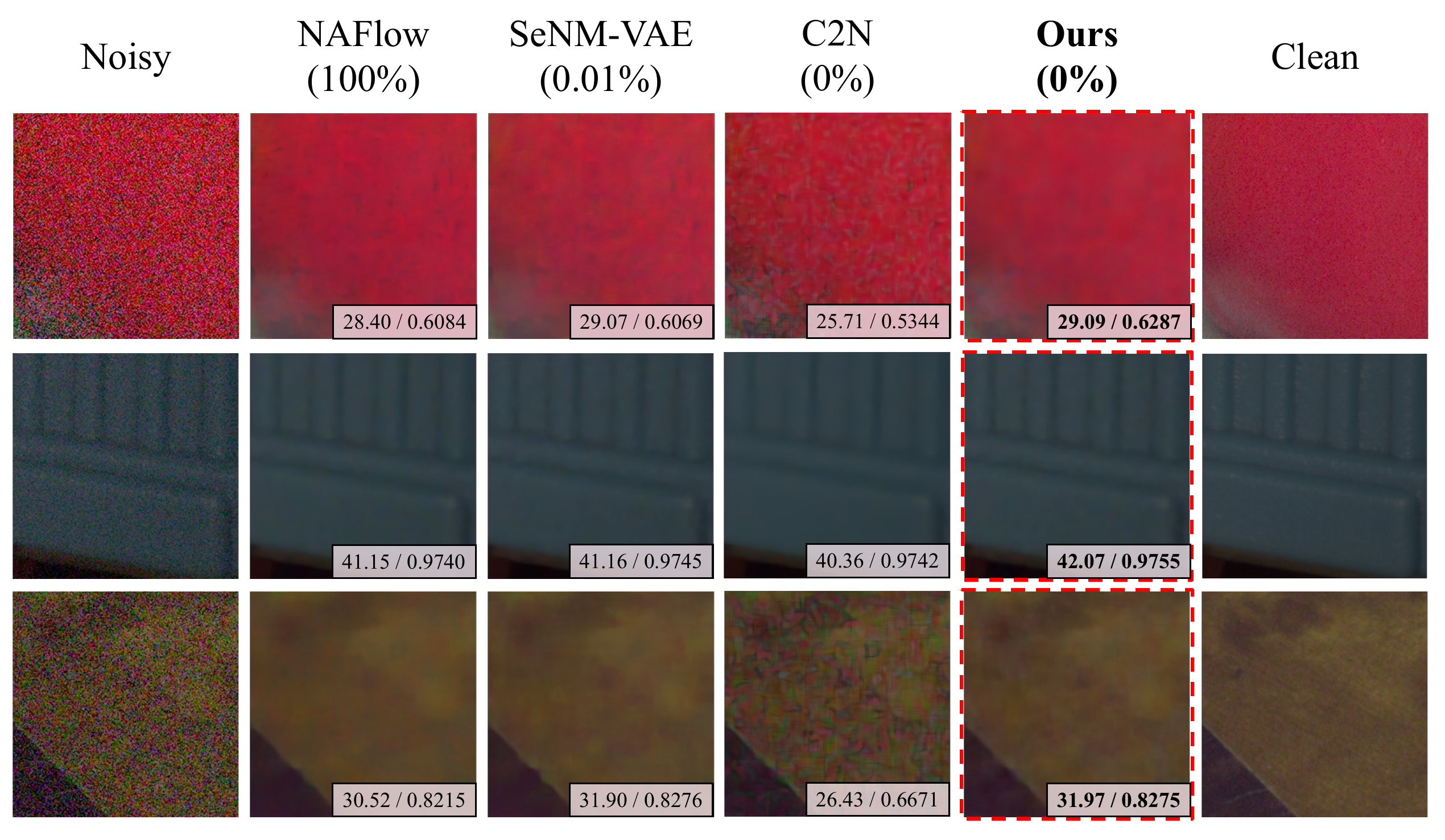}}
\caption{Visualization of AP-BSN denoising results on the SIDD validation set. Numbers below each image denote PSNR / SSIM.}
\vspace{-12mm}
\label{fig:apbsn_result_supple}
\end{center}
\end{figure*}

\begin{figure*}[t]
\begin{center}
\centerline{\includegraphics[width=0.9\textwidth]{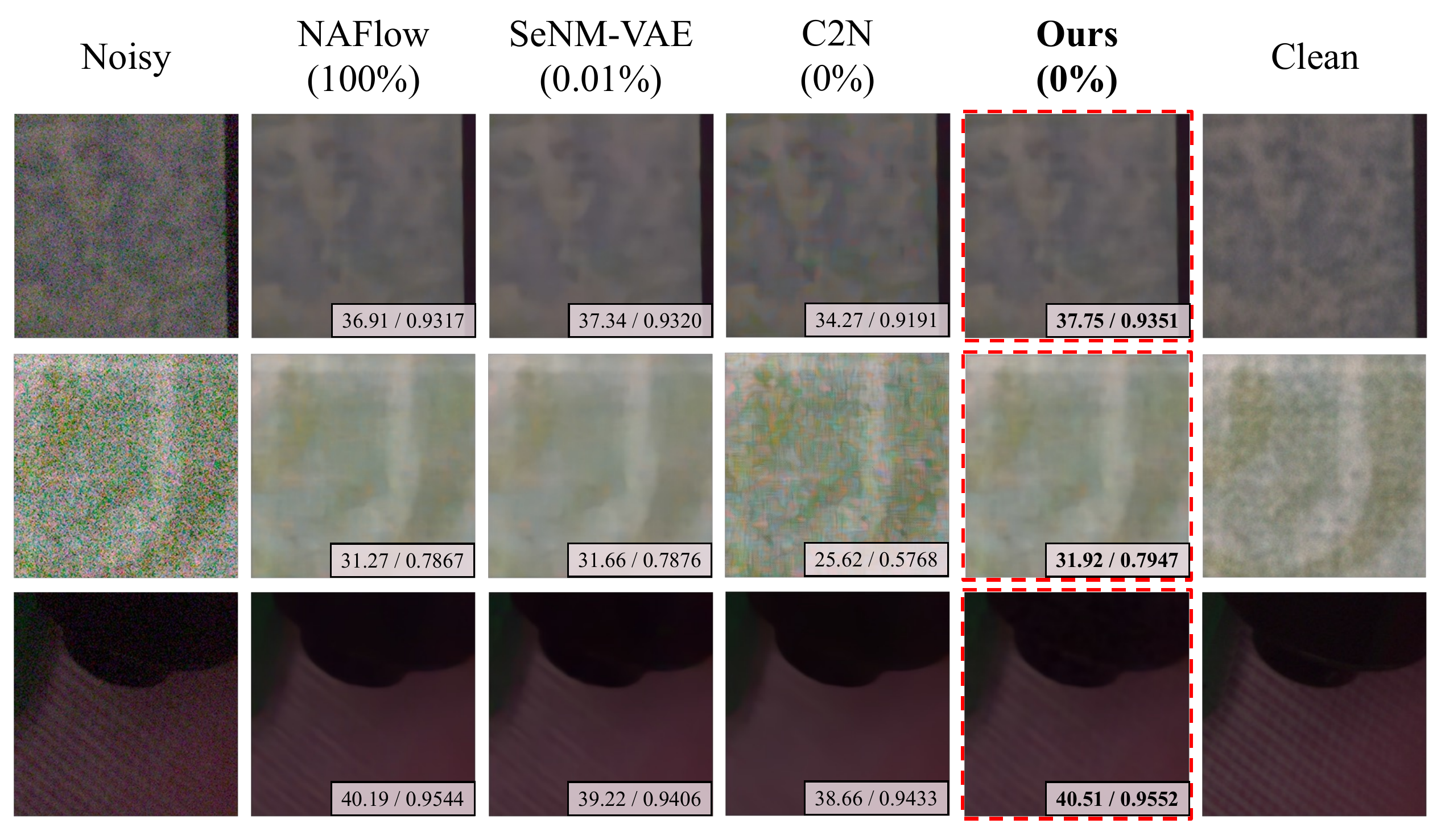}}
\caption{Visualization of MM-BSN denoising results on the SIDD validation set. Numbers below each image denote PSNR / SSIM.}
\vspace{-12mm}
\label{fig:mmbsn_result_supple}
\end{center}
\end{figure*}

In \ffigref{fig:apbsn_result_supple} and \ffigref{fig:mmbsn_result_supple}, we provide additional visualizations of denoising results, where the denoising network is trained on synthetic datasets. We compare our method (\framework{}) with other approaches, including NAFlow, SeNM-VAE, and C2N.
Visually, denoisers trained on C2N data struggle to completely remove real-world noise and tend to leave noticeable visual artifacts. While models trained on NAFlow and SeNM-VAE successfully remove most of the noise, they exhibit a tendency to over-smooth the images, occasionally compromising intrinsic structural patterns and fine textures present in the clean reference. Conversely, denoisers trained on the dataset synthesized by \framework{} effectively remove complex sRGB noise while preserving the original structural patterns and sharp edges. This demonstrates that the diverse and realistic noise priors generated by our method enhance the robustness of downstream denoising tasks without sacrificing scene fidelity.

\vspace{-2mm}
\section{Limitations}
\label{suppl_limit}

The proposed \framework{} framework operates on the assumption of structural consistency across the input burst, requiring two consecutively captured noisy images of a static scene. As demonstrated in Section~\ref{suppl_misalignment}, our method exhibits inherent robustness to moderate spatial misalignments. However, in scenarios where the geometric discrepancy becomes excessively severe, the subtractive injection mechanism may encounter large-scale feature misalignment. Such severe inconsistencies could potentially exceed the model's capacity for robust disentanglement, leading to suboptimal separation of noise and scene content within the Reconstruction Autoencoder.
Although utilizing burst noisy pairs relaxes the strict requirement for clean references or metadata, it still restricts the applicability of the model in highly dynamic environments. To address this limitation, future work will focus on incorporating explicit feature-level alignment modules to handle complex motions. Furthermore, we plan to extend the generative framework to operate on a single noisy image or within a completely unpaired setting, thereby entirely eliminating the dependency on burst acquisitions.

\clearpage

%
%


\end{document}